\documentclass[runningheads]{llncs}

\usepackage{eccv}

\usepackage{eccvabbrv}

\usepackage{graphicx}
\usepackage{booktabs}
\usepackage{subcaption}
\usepackage{color, colortbl}
\usepackage{xcolor}
\usepackage{multirow}
\usepackage[utf8]{inputenc}
\usepackage{csquotes}
\usepackage{array}
\usepackage{multirow}
\usepackage{siunitx}
\usepackage[normalem]{ulem}
\usepackage{soul}
\usepackage{amssymb}
\usepackage{pifont}
\usepackage[hang,flushmargin]{footmisc} 

\usepackage[accsupp]{axessibility}  

\usepackage{caption}

\usepackage[pagebackref=true,breaklinks=true,letterpaper=true,colorlinks,bookmarks=false]{hyperref}

\usepackage{orcidlink}
\NewDocumentCommand\emojiclip{}{\raisebox{-0.5mm}{\includegraphics[scale=0.016]{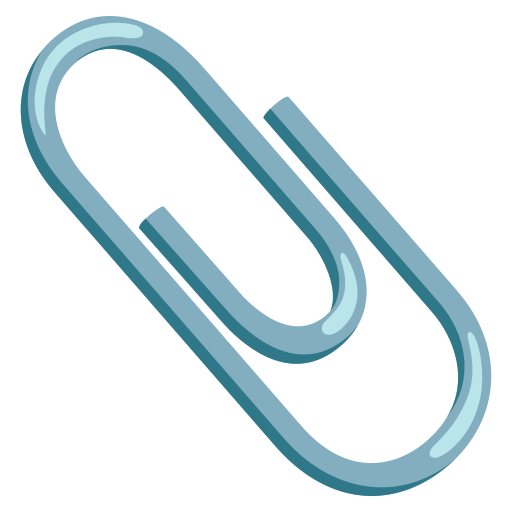}}}
\NewDocumentCommand\emojidino{}{\raisebox{-0.5mm}{\includegraphics[scale=0.017]{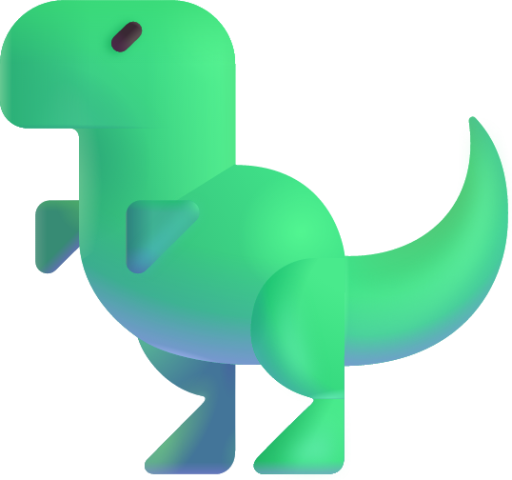}}}
\NewDocumentCommand\emojieva{}{\raisebox{-0.5mm}{\includegraphics[scale=0.004]{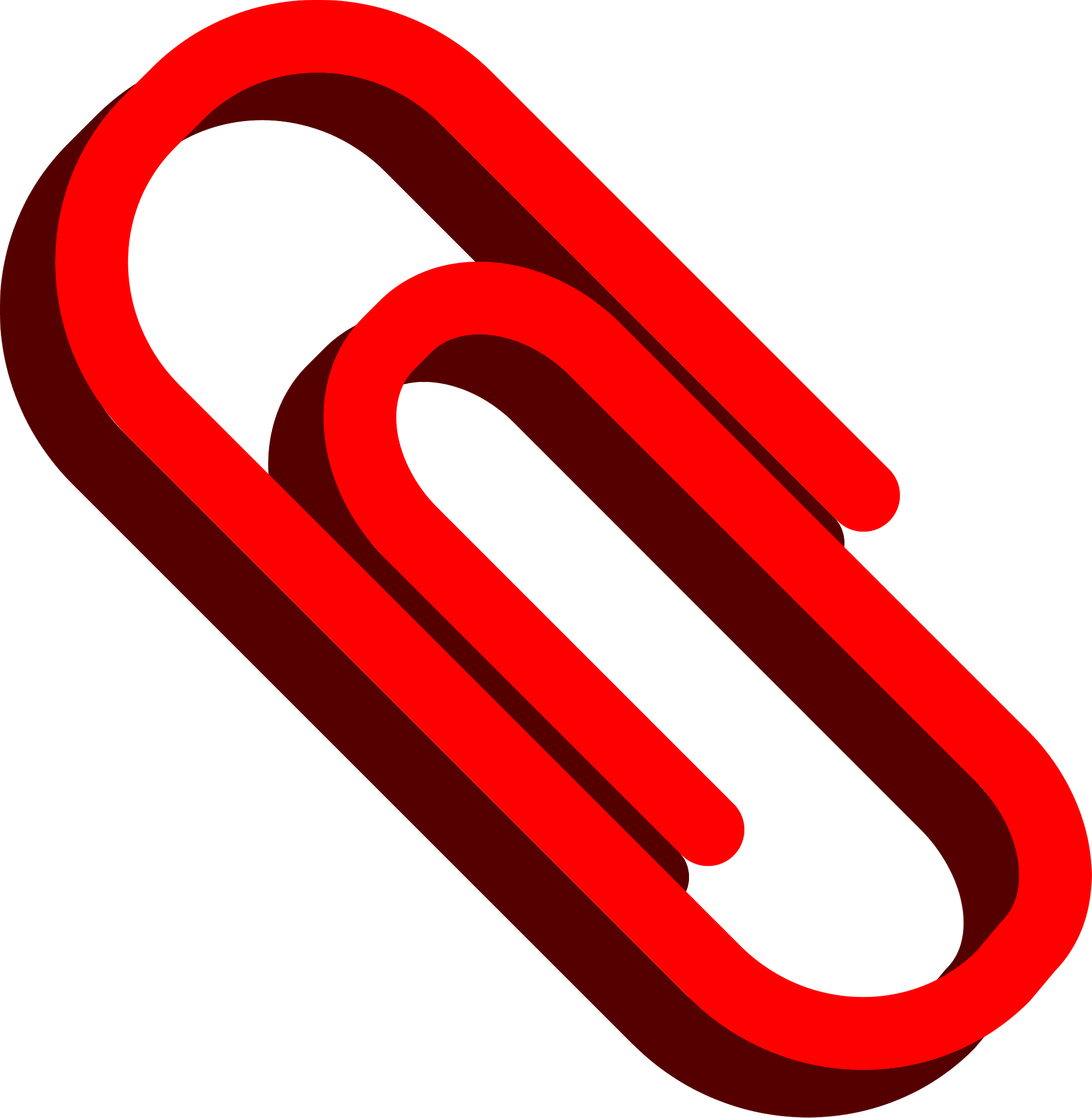}}}
\NewDocumentCommand\emojicross{}{\raisebox{-0.4mm}{\includegraphics[scale=0.033]{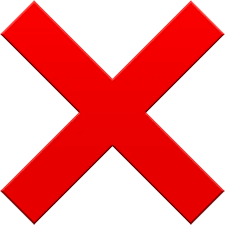}}}
\definecolor{qrand}{HTML}{2b2e4a}
\definecolor{qtext}{HTML}{e84545}
\definecolor{blue}{HTML}{e8f0f8}

\begin{document}

\title{Textual Query-Driven Mask Transformer {\\} for Domain Generalized Segmentation} 

\titlerunning{Textual Query-Driven Mask Transformer}

\author{\hspace{4.5mm}  Byeonghyun Pak$^\ast$
\quad \quad
Byeongju Woo$^\ast$
\quad \quad
Sunghwan Kim$^\ast$
\quad \quad \vspace{1.2mm} {\\}
Dae-hwan Kim 
\quad \quad
Hoseong Kim$^\dagger$}

\authorrunning{B.~Pak et al.}

\institute{Agency for Defense Development (ADD)}

\maketitle

\begin{abstract}
  In this paper, we introduce a method to tackle Domain Generalized Semantic Segmentation (DGSS) by utilizing domain-invariant semantic knowledge from text embeddings of vision-language models. We employ the text embeddings as object queries within a transformer-based segmentation framework (textual object queries). These queries are regarded as a domain-invariant basis for pixel grouping in DGSS. To leverage the power of textual object queries, we introduce a novel framework named the {\large\texttt{t}}extual {\large\texttt{q}}uery-{\large\texttt{d}}riven {\large\texttt{m}}ask transformer (\texttt{tqdm}). Our \texttt{tqdm} aims to (1) generate textual object queries that maximally encode domain-invariant semantics and (2) enhance the semantic clarity of dense visual features. Additionally, we suggest three regularization losses to improve the efficacy of \texttt{tqdm} by aligning between visual and textual features. By utilizing our method, the model can comprehend inherent semantic information for classes of interest, enabling it to generalize to extreme domains (\eg, sketch style). Our \texttt{tqdm} achieves 68.9 mIoU on GTA5$\rightarrow$Cityscapes, outperforming the prior state-of-the-art method by 2.5 mIoU. 
  The project page is available at \url{https://byeonghyunpak.github.io/tqdm}.

\end{abstract}

\def\thefootnote{*}\footnotetext{Equal contribution \quad $^\dagger$Corresponding author}
\def\thefootnote{\arabic{footnote}}



\section{Introduction}
Developing a model that generalizes robustly to unseen domains has been a long-standing goal in the field of machine perception. In this context, Domain Generalized Semantic Segmentation (DGSS) aims to build models that can effectively operate across diverse target domains, trained solely on a single source domain. This area has made notable improvements with a wide range of approaches, including normalization and whitening \cite{choi2021robustnet, pan2018two, peng2022semantic, pan2019switchable}, domain randomization \cite{kim2023texture, lee2022wildnet, yue2019domain, zhao2022style, fahes2023simple}, and utilizing the inherent robustness of transformers \cite{ding2023hgformer, sun2023ibaformer}. 

\begin{figure}[t]
    \centering
    \begin{subfigure}[t]{0.3\linewidth}
        \centering
        \phantomsubcaption{}
        \label{fig:teaser_a}
    \end{subfigure}
    \begin{subfigure}[t]{0.3\linewidth}
        \centering
        \phantomsubcaption{}
        \label{fig:teaser_b}
    \end{subfigure}
    \begin{subfigure}[t]{0.3\linewidth}
        \centering
        \phantomsubcaption{}
        \label{fig:teaser_c}
    \end{subfigure}
    \includegraphics[width=\linewidth]{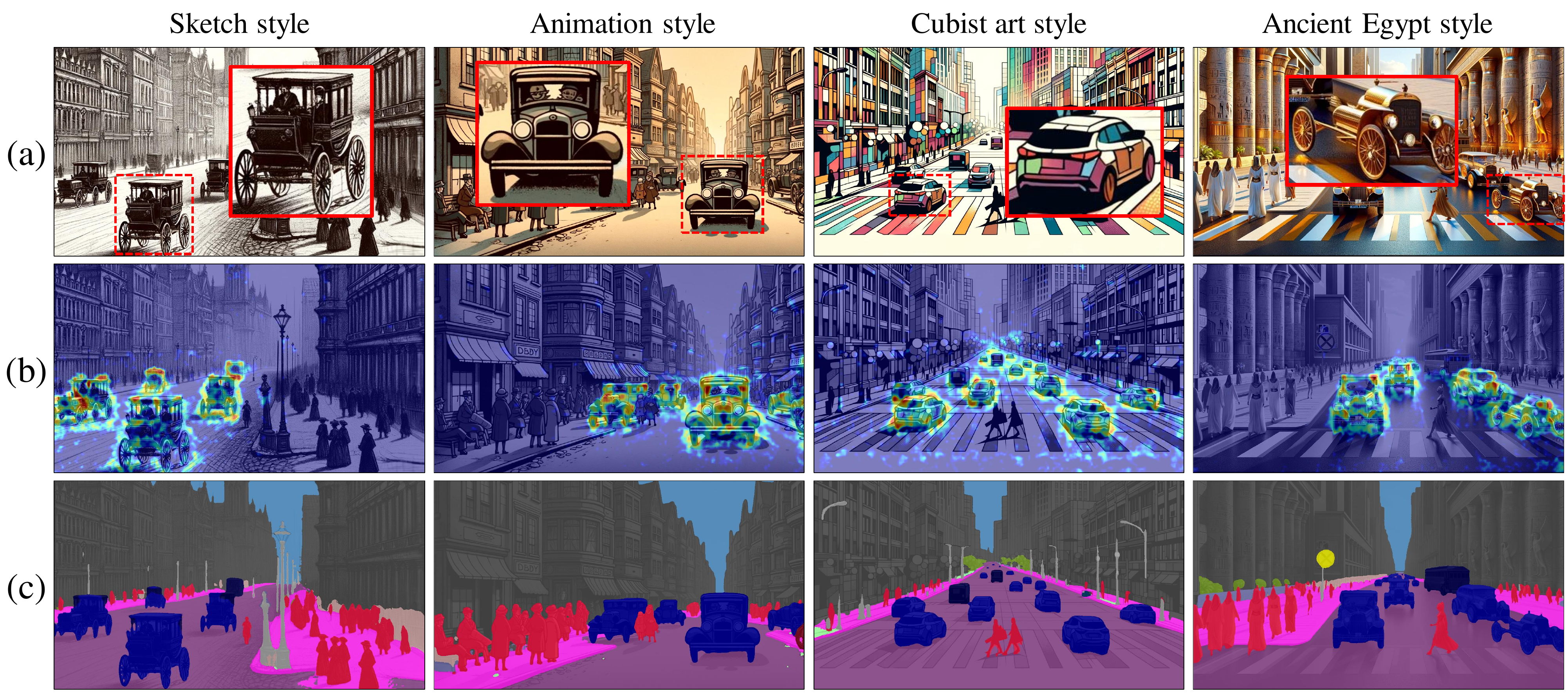}
    \caption{(a) A collection of driving scene images with diverse styles generated by ChatGPT\protect\footref{fn:gpt}. {(b) The image-text similarity maps of a pre-trained VLM (\ie, EVA02-CLIP \cite{sun2023eva}) on diverse domains. The text embedding of `car' is consistently well-aligned with the corresponding class regions of images across various domains.} (c) The segmentation results predicted by our proposed \texttt{tqdm}.
    {Note that our model can generalize to extreme domains (\eg, sketch style) and effectively identify the cars in multiple forms that are not present in the source domain (\ie, GTA5 \cite{richter2016playing}).}}
    \label{fig:teaser}
    \vspace{-4mm}
\end{figure}

Meanwhile, the recent advent of Vision-Language Models (VLMs) (\eg, CLIP \cite{radford2021learning}) has introduced new possibilities and applications in various vision tasks, thanks to their rich semantic representations \cite{zhu2023survey}. One notable strength of VLMs is their ability to generalize across varied domain shifts \cite{radford2021learning, nguyen2022quality}. This capability has inspired the development of methods for domain generalization in image classification \cite{mangla2022indigo,huang2023sentence,wortsman2022robust,cho2023promptstyler}. Furthermore, there have been efforts to incorporate the robust visual representation of VLMs in DGSS\cite{hummer2023vltseg,fahes2023simple}. 
However, existing methods in DGSS have not yet explored direct language-driven approaches to utilize textual representations from VLMs for domain-generalized recognition.
Note that the contrastive learning objective in VLMs aligns a text caption (\eg, \texttt{``a photo of a car''}) with images from a wide range of domains in a joint space \cite{radford2021learning}. This alignment enables the text embeddings to capture domain-invariant semantic knowledge \cite{huang2023sentence}, as demonstrated in \cref{fig:teaser}

In this paper, we introduce a language-driven DGSS method that harnesses domain-invariant semantics from textual representations in VLMs. Our proposed method can make accurate predictions even under extreme domain shifts, as it comprehends the inherent semantics of targeted classes. In \cref{fig:teaser_c}, our model exhibits accurate segmentation results on driving scene images with diverse styles generated by ChatGPT\footnote{\url{https://chat.openai.com}\label{fn:gpt}}. Note that our model can effectively identify cars in multiple forms that are not present in the source domain (\ie, GTA5 \cite{richter2016playing}).

The key idea of our method is to utilize text embeddings of classes of interest from VLMs as object queries, referred to as \textit{textual object queries}. Given that an object query can be considered as a {mask embedding} vector to group regions belonging to the same class \cite{cheng2022masked}, textual object queries generate robust {mask predictions} for classes of interest across diverse domains, thanks to their domain-invariant semantics (\cref{sec-3}). Building on this insight, we propose a {\large\texttt{t}}extual {\large\texttt{q}}uery-{\large\texttt{d}}riven {\large\texttt{m}}ask transformer (\texttt{tqdm}) to leverage textual object queries for DGSS.
Our design philosophy lies in (1) generating the queries that maximally encode domain-invariant semantic knowledge and (2) enhancing their adaptability in dense predictions by improving the semantic clarity of pixel features (\cref{sec-4.1}). Additionally, we discuss three regularization losses to preserve the robust vision-language alignment of pre-trained VLMs, thereby improving the effectiveness of our method (\cref{sec-4.2}). 

Our contributions can be summarized into three aspects:
\begin{itemize}
    \item To the best of our knowledge, we are the first to introduce a direct language-driven DGSS approach using text embeddings from VLMs to enable domain-invariant recognition, effectively handling extreme domain shifts.
    \item We propose a novel query-based segmentation framework named \texttt{tqdm} that leverages textual object queries, along with three regularization losses to support this framework.
    \item Our \texttt{tqdm} demonstrates the state-of-the-art performance across multiple DGSS benchmarks; \eg, \texttt{tqdm} achieves 68.9 mIoU on GTA5$\rightarrow$Cityscapes, which improves the prior state-of-the-art method by 2.5 mIoU.
\end{itemize}


\section{Related Work}
\noindent \textbf{Vision-Language Models (VLMs).} VLMs \cite{tan2019lxmert,desai2021virtex,lu2019vilbert,radford2021learning,jia2021scaling,pham2023combined}, which are trained on extensive web-based image-caption datasets, have gained attention for their rich semantic understanding. CLIP \cite{radford2021learning} employs contrastive language-image pre-training, and several studies \cite{radford2021learning, nguyen2022quality} have explored its robustness to natural distribution shifts. EVA02-CLIP\cite{fang2023eva,fang2023eva02} further enhances CLIP by exploring robust dense visual features through masked image modeling\cite{he2022masked}.

The advanced capabilities of VLMs enable more challenging segmentation tasks. For example, open-vocabulary segmentation \cite{li2022languagedriven, xu2022simple, liang2023open,zhou2023zegclip} attempts to segment an image by arbitrary categories described in texts. Although these works use text embeddings for segmentation tasks similar to our approach, our work distinctly differs in the problem of interest and the solution. We aim to build models that generalize well to unseen domains, whereas the prior works primarily focus on adapting to unseen classes.

\vspace{2mm}
\noindent \textbf{Domain Generalized Semantic Segmentation (DGSS).} DGSS aims to develop a robust segmentation model that can generalize robustly across various unseen domains. Prior works have concentrated on learning domain-invariant representations through approaches such as normalization and whitening \cite{pan2018two, choi2021robustnet, peng2022semantic, pan2019switchable}, and domain randomization \cite{lee2022wildnet, zhao2022style, zhao2023style, kim2023texture, yue2019domain}. Normalization and whitening remove domain-specific features to focus on learning domain-invariant features. For instance, RobustNet \cite{choi2021robustnet} selectively whitens features sensitive to photometric changes. Domain randomization seeks to diversify source domain images by augmenting them into various domain styles. DRPC \cite{yue2019domain} ensures consistency among multiple stylized images derived from a single content image. TLDR \cite{kim2023texture} explores domain randomization while focusing on learning textures. The incorporation of vision transformers \cite{ding2023hgformer, sun2023ibaformer} has further enhanced DGSS by utilizing the robustness of attention mechanisms. However, these methods largely correspond to visual pattern recognition and
have limited ability in understanding high-level semantic concepts inherent to each class.

Recent studies \cite{fahes2023simple, hummer2023vltseg, wei2023stronger} have attempted to utilize VLMs in DGSS. Rein \cite{wei2023stronger} introduces an efficient fine-tuning method that preserves the generalization capability of large-scale vision models, including CLIP\cite{radford2021learning} and EVA02-CLIP\cite{fang2023eva,fang2023eva02}. FAMix \cite{fahes2023simple} employs language as the source of style augmentation, along with a minimal fine-tuning method for the vision backbone of VLMs.
VLTseg\cite{hummer2023vltseg} fine-tunes the vision encoder of VLMs, aligning dense visual features with text embeddings via auxiliary loss. Despite these advancements, the existing approaches either do not utilize language information \cite{wei2023stronger} or use it primarily as an auxiliary tool to support training pipelines \cite{fahes2023simple, hummer2023vltseg}. In contrast, this paper introduces a direct language-driven approach to fully harness domain-invariant semantic information embedded in the textual features of VLMs.

\vspace{2mm}
\noindent \textbf{Object query design.} Recent studies\cite{zhang2022segvit,cheng2021per,cheng2022masked,yu2022k,li2022panoptic} have explored query-based frameworks that utilize a transformer decoder\cite{vaswani2017attention} for segmentation tasks, inspired by DETR \cite{carion2020end}. 
In these frameworks, an object query serves to group pixels belonging to the same semantic region (\eg, object or class) by representing the region as a latent vector. Given the critical role of object queries, existing studies have focused on their design and optimization strategies \cite{kamath2021mdetr,liu2022dab,li2022dn,zhang2022dino,cheng2022masked,liu2023boosting, zhang2023mp,cai2022x}. 
Mask2former\cite{cheng2022masked} guides query optimization via masked cross-attention for restricting the query to focus on predicted segments. ECENet\cite{liu2023boosting} generates object queries from predicted masks to ensure that the queries represent explicit semantic information. Furthermore, several works\cite{li2022dn,zhang2023mp,zhang2022dino} have introduced additional supervision into object queries to enhance training stability, 
{while others\cite{kamath2021mdetr,cai2022x} have designed conditional object queries for cross-modal tasks.}

While these studies have demonstrated that purpose-specific object queries contribute to performance, convergence, and functionality in dense prediction tasks\cite{li2023transformer}, research on object queries for DGSS remains unexplored. Our work aims to address DGSS by designing domain-invariant object queries and developing a decoder framework to improve the adaptability of these object queries.


\section{Textual Object Query} \label{sec-3}
Our observation is that utilizing text embeddings from VLMs as object queries within a query-based segmentation framework enables domain-invariant recognition. This recognition leads to the effective grouping of dense visual features across diverse domains.

Recent studies \cite{huang2023sentence,mangla2022indigo} have suggested that the text embedding of a class captures core semantic concepts to represent the class across different visual domains, \ie, domain invariant semantics. This capability stems from web-scale contrastive learning\cite{radford2021learning}, which aligns the text embedding for a class of interest with image features of the corresponding class from a wide variety of domains. Given that the text embeddings have the potential to align with dense visual features\cite{rao2022denseclip,zhou2022extract,mukhoti2023open},
one can leverage the textual information for domain-generalized dense predictions. We find that the text embedding of a class name is well-aligned with the visual features of the class region, even under extreme domain shifts (see \cref{sec:A}).

\begin{figure}[t]
    \centering
    \begin{subfigure}[t]{0.49\linewidth}
        \centering
        \phantomsubcaption{}
        \label{fig:textual_object_query_a}
    \end{subfigure}
    \begin{subfigure}[t]{0.49\linewidth}
        \centering
        \phantomsubcaption{}
        \label{fig:textual_object_query_b}
    \end{subfigure}
    \includegraphics[width=1\linewidth]{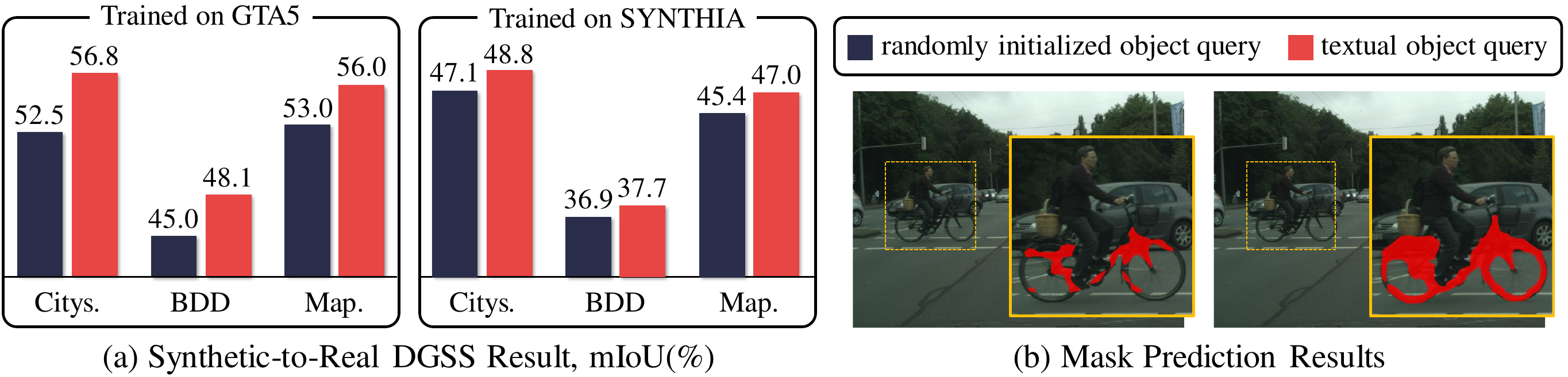}
    \caption{\textbf{Effectiveness of textual object query.} 
    {
    (a) In all DGSS benchmarks, textual object queries (\textcolor{qtext}{$\textbf{q}_\text{text}$}) outperforms randomly initialized object queries (\textcolor{qrand}{$\textbf{q}_\text{rand}$}).}
    (b) We visualize the mask predictions corresponding to a class (\ie, `bicycle'), derived from $\textbf{q}_\text{rand}$ on the left and $\textbf{q}_\text{text}$ on the right, respectively. $\textbf{q}_\text{rand}$ yields a degraded result, whereas $\textbf{q}_\text{text}$ produces a robust one on a unseen domain.}
    \label{fig:textual_object_query}
    \vspace{-5mm}
\end{figure}

To leverage domain-invariant semantics in textual features from VLMs, we propose utilizing \textit{textual object queries}.
Generally, object queries in transformer-based segmentation frameworks are conceptualized as mask embedding vectors, representing regions likely to be an `object' or a `class' \cite{cheng2022masked}. In semantic segmentation, the queries are optimized to represent the semantic information for classes of interest. Our key insight is that designing object queries with generalized semantic information for classes of interest leads to domain-invariant recognition. We implement textual object queries using the text embeddings of targeted classes from the text encoder of VLMs.

We conduct a motivating experiment to demonstrate the capability of textual object queries to generalize to unseen domains. We design a simple architecture comprising an encoder and textual object queries ($\textbf{q}_\text{text}$), and compare it with an architecture that employs conventional, randomly initialized object queries ($\textbf{q}_\text{rand}$). The details of this experiment are described in \cref{sec:B}. In \cref{fig:textual_object_query_a}, $\textbf{q}_\text{text}$ outperforms $\textbf{q}_\text{rand}$ in all unseen target domains. \cref{fig:textual_object_query_b} clearly supports this observation: $\textbf{q}_\text{rand}$ yields a degraded mask prediction for a `bicycle,' whereas $\textbf{q}_\text{text}$ produces a robust result on a unseen domain. {\large\texttt{q}}uery {\large\texttt{d}}riven {\large\texttt{m}}ask transformer (\texttt{tqdm}) to leverage the power of textual object queries.


\section{Method}
\begin{figure}[t]
    \centering
    \includegraphics[width=1\linewidth]{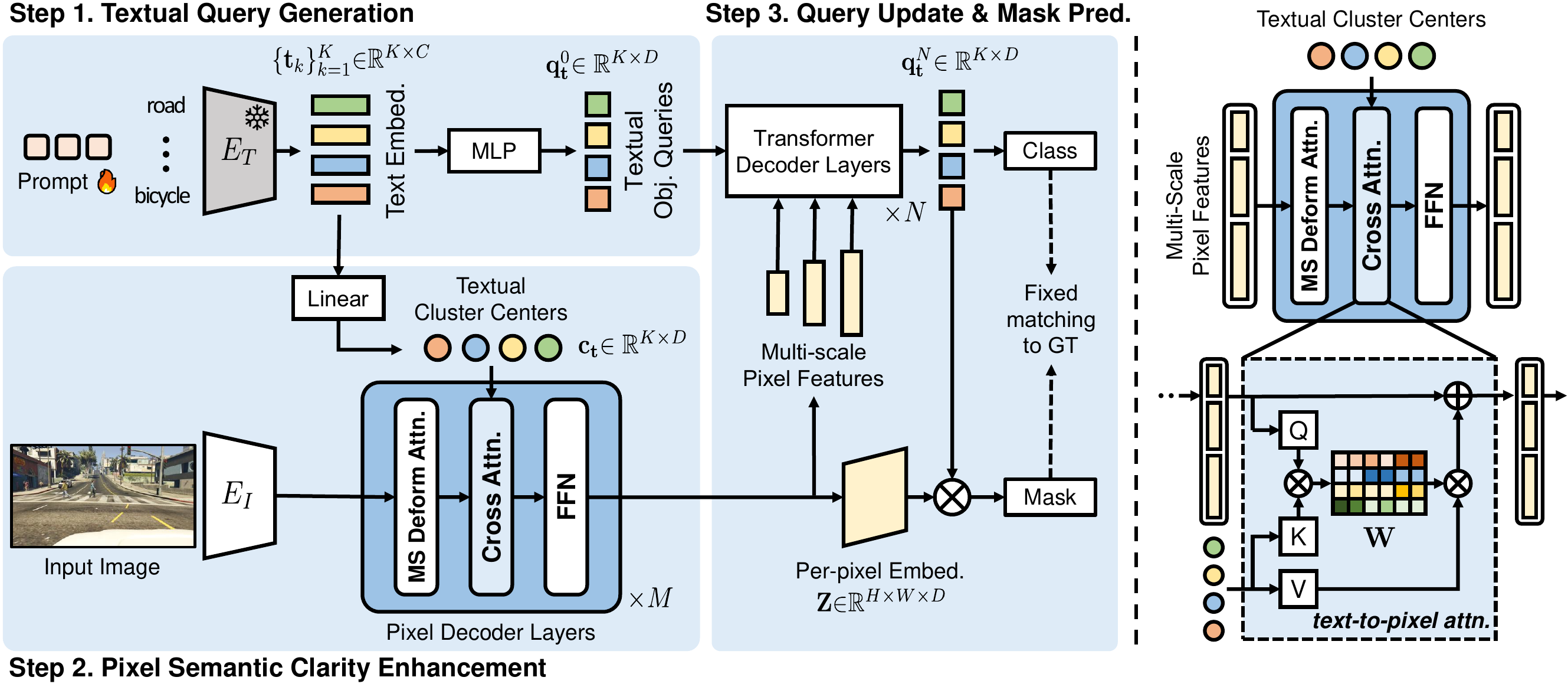}
    \caption{\textbf{Overall pipeline of \texttt{tqdm}.} (Step 1) We generate initial textual object queries $\textbf{q}^0_\textbf{t}$ from the $K$ class text embeddings $\{\textbf{t}_k\}^K_{k=1}$. (Step 2) To improve the segmentation capabilities of these queries, we incorporate text-to-pixel attention within the pixel decoder. This process enhances the semantic clarity of pixel features, while reconstructing high-resolution per-pixel embeddings $\textbf{Z}$. (Step 3) The transformer decoder refines these queries for the final prediction. Each prediction output is then assigned to its corresponding ground truth (GT) through fixed matching, ensuring that each query consistently represents the semantic information of one class.}
    \label{fig:overview}
    \vspace{-5mm}
\end{figure}
In this section, we propose {\large\texttt{t}}extual {\large\texttt{q}}uery-{\large\texttt{d}}riven {\large\texttt{m}}ask transformer (\texttt{tqdm}), a segmentation decoder that comprehends domain-invariant semantic knowledge by leveraging \textit{textual object queries} as pixel grouping basis (\cref{sec-4.1}). Furthermore, we discuss three regularization losses that aim to maintain 
{robust vision-language alignment,}
thereby enhancing the efficacy of \texttt{tqdm} (\cref{sec-4.2}). 

\vspace{2mm}
\noindent \textbf{Preliminary.} 
We employ the image encoder $E_I$ and text encoder $E_T$, both of which are initialized with a pre-trained Vision-Language Model (VLM) (\eg, CLIP\cite{radford2021learning}). We fully fine-tune $E_I$ to learn enhanced dense visual representations, whereas $E_T$ is kept frozen to preserve robust textual representations. $E_I$ extracts multi-scale pixel features from an image and feeds them into our \texttt{tqdm} decoder. Additionally, $E_I$ outputs visual embeddings $\textbf{x}$, which are projected in a joint vision-language space.

\subsection{Textual Query-Driven Mask Transformer} \label{sec-4.1}

Our proposed framework, \texttt{tqdm}, aims to leverage textual object queries for DGSS in three key steps. Initially, \textit{textual query generation} focuses on generating textual object queries that maximally encode domain-invariant semantic knowledge. Subsequently, \textit{pixel semantic clarity enhancement} aims to improve the segmentation capability of textual object queries by incorporating text-to-pixel attention within a pixel decoder. Lastly, following the practices of mask transformer\cite{cheng2021per,cheng2022masked}, a transformer decoder refines the object queries for the final mask prediction.
The overall pipeline of \texttt{tqdm} is demonstrated in \cref{fig:overview}.

\vspace{2mm}
\noindent \textbf{Textual query generation.}
To generate textual object queries for DGSS, we prioritize two key aspects: the queries should (1) preserve domain-invariant semantic information for robust prediction and (2) adapt to the segmentation task to ensure promising performance. To meet the first requirement, we maintain $E_T$ frozen to preserve its original language representations. For the second requirement, we employ learnable prompts \cite{zhou2022learning} to adapt textual features from $E_T$. 
Specifically, $E_T$ generates a text embedding $\textbf{t}_k$$\in$$\mathbb{R}^{C}$ for each class label name embedding $\{\text{class}_k\}$ with a learnable prompt $\textbf{p}$: 
\begin{equation}
    \textbf{t}_k=E_T([\textbf{p},\{\text{class}_k\}]),
\end{equation}
where $1$$\leq$$k$$\leq$$K$ for total $K$ classes, and $C$ denotes the channel dimension.
Finally, we obtain initial textual object queries $\textbf{q}_\textbf{t}^0 \in\mathbb{R}^{K\times D}$ from the text embeddings {$\textbf{t}$$=$$\{\textbf{t}_k\}^K_{k=1}$$\in$$\mathbb{R}^{K\times C}$} through multi-layer perceptron (MLP). Note that $D$ is the dimension of the query vectors in \texttt{tqdm}.

\vspace{2mm}
\noindent \textbf{Pixel semantic clarity enhancement.} 
To improve the segmentation capabilities of textual object queries, we incorporate a \textit{text-to-pixel attention} mechanism that enhances the semantic clarity of each pixel feature (refer to ``Cross Attn.'' in \cref{fig:overview}).
This mechanism ensures that pixel features are clearly represented in terms of domain-invariant semantics, allowing them to be effectively grouped by textual object queries.

Initially, we derive textual cluster centers $\textbf{c}_\textbf{t}$$\in$$\mathbb{R}^{K\times D}$ by compressing the channel dimension of the text embeddings 
{$\textbf{t}$$\in$$\mathbb{R}^{K\times C}$}
with a linear layer. Then, within a pixel decoder layer, a text-to-pixel attention block utilizes multi-scale pixel features $\textbf{z}$$\in$$\mathbb{R}^{L\times D}$ as query tokens $\textbf{Q}_{\textbf{z}}$ with a linear projection. Here, $L$ denotes the length of pixel features. The textual clustering centers $\textbf{c}_\textbf{t}$ are projected into key tokens $\textbf{K}_{\textbf{t}}$ and value tokens $\textbf{V}_{\textbf{t}}$. The attention weights $\textbf{W}$$\in$$\mathbb{R}^{L\times K}$ and the enhanced pixel features are calculated as follows:
\begin{align}
	\textbf{W} &= \operatorname{softmax}(\textbf{Q}_{\textbf{z}}\textbf{K}_\textbf{t}^\top),\\
	\textbf{z} &\leftarrow \textbf{z} + \textbf{W}\textbf{V}_\textbf{t}. \label{eq:z'}
\end{align}
We consider this text-to-pixel attention mechanism as a method for updating the pixel features toward $K$ textual clustering centers.
The attention weight $\textbf{W}$ calculates similarity scores between the $L$ pixel features and the $K$ textual clustering centers, where $\textbf{K}_\textbf{t}$ serves as the clustering centers. Then, in \cref{eq:z'}, we refine the pixel features $\textbf{z}$ with $\textbf{W}\textbf{V}_\textbf{t}$, aiming to align more closely with the  $K$ textual clustering centers. This approach promotes the grouping of regions belonging to the same classes, thereby improving their semantic clarity.

\vspace{2mm}
\noindent \textbf{Query update and mask prediction.}
The final step involves updating textual object queries and predicting segmentation masks. The transformer decoder, including masked attention\cite{cheng2022masked} with $N$ layers, progressively refines the initial textual object queries $\textbf{q}_{\textbf{t}}^0$ into $\textbf{q}_{\textbf{t}}^N$ by integrating pixel features from the pixel decoder. 
The refined textual object queries $\textbf{q}_{\textbf{t}}^N$$\in$$\mathbb{R}^{K\times D}$ then predict $K$ masks via dot product with per-pixel embeddings $\textbf{Z}$$\in$$\mathbb{R}^{H\times W\times D}$ followed by sigmoid activation, where $H$ and $W$ are the spatial resolutions.
These queries are then classified by a linear classifier with softmax activation to produce a set of class probabilities.
We optimize \texttt{tqdm} using the segmentation loss $\mathcal{L}_{\mathrm{seg}}$, following \cite{cheng2022masked}:
\begin{equation}
 \mathcal{L}_{\mathrm{seg}} = 
    \lambda_{\mathrm{bce}}\mathcal{L}_{\mathrm{bce}} + 
    \lambda_{\mathrm{dice}}\mathcal{L}_{\mathrm{dice}} +
    \lambda_{\mathrm{cls}}\mathcal{L}_{\mathrm{cls}},
\end{equation}
where the binary cross-entropy loss $\mathcal{L}_{\mathrm{bce}}$ and the dice loss \cite{milletari2016v} $\mathcal{L}_{\mathrm{dice}}$ optimize the predicted masks, and the categorical cross-entropy loss $\mathcal{L}_{\mathrm{cls}}$ optimizes the class prediction of queries. The loss weights $\lambda_{\mathrm{bce}}$, $\lambda_{\mathrm{dice}}$ and $\lambda_{\mathrm{cls}}$ are set to the same values as those in \cite{cheng2022masked}. 
To assign each query to a specific class, we adopt fixed matching instead of bipartite matching (see \cref{tab:ablation-c} for ablation). This matching ensures that each query solely represents the semantic information of one class.
\begin{figure}[!t]
    \centering
    \includegraphics[width=0.93\linewidth]{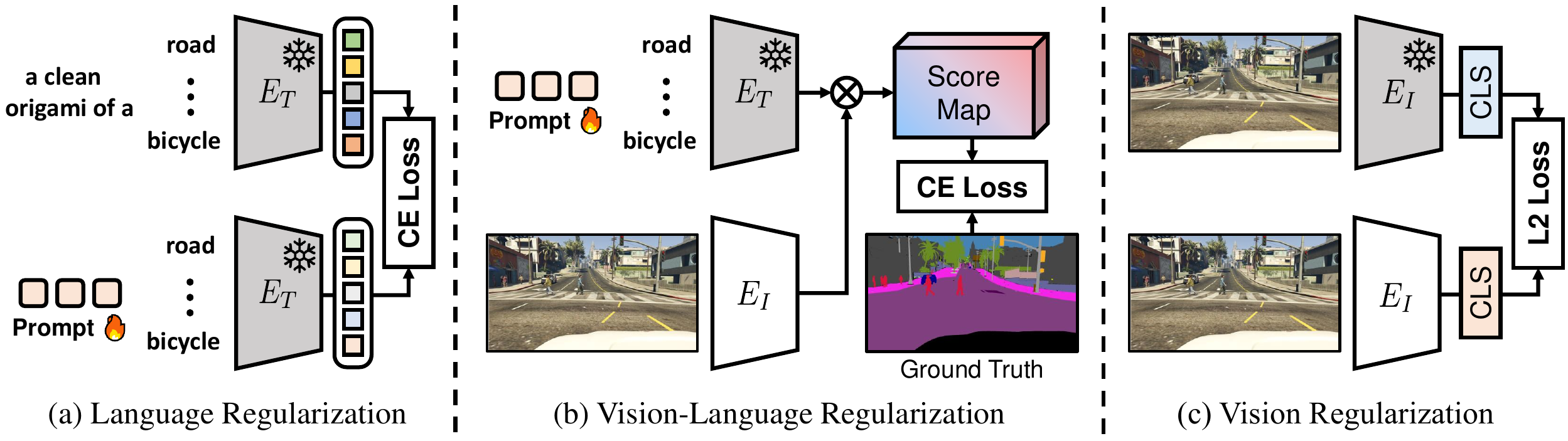}
    \vspace{-1mm}
    \caption{\textbf{Three regularization losses to enhance the efficacy of \texttt{tqdm}.} (a) Language regularization prevents the learnable prompts from distorting the semantic meaning of text embeddings. (b) Vision-language regularization aims to align visual and textual features at the pixel-level. (c) Vision regularization maintains the ability of the vision encoder to align with textual information at the image-level.}
    \label{fig:fig-reg}
    \vspace{-5mm}
\end{figure}

\subsection{Regularization} \label{sec-4.2}
Our textual query-driven approach is based on a strong alignment between visual and textual features. To maintain this alignment, we propose three regularization strategies: (1) language regularization that prevents the learnable prompts from distorting the semantic meaning of text embeddings, (2) vision-language regularization that ensures the pixel-level alignment between visual and textual features, and (3) vision regularization that preserves the textual alignment capability of the vision encoder from a pre-trained VLM. 
The regularization losses are demonstrated in \cref{fig:fig-reg}. 

\vspace{2mm}
\noindent\textbf{Language regularization.} 
We use learnable text prompts \cite{zhou2022learning} to adapt text embeddings for DGSS. During training, the prompts can distort the semantic meaning of text embeddings. To address this issue, we introduce a language regularization loss, which ensures semantic consistency between the text embeddings derived from the learnable prompt $\textbf{p}$ and those derived from the fixed prompt $\textbf{P}_0$. These embeddings are denoted as $\textbf{t}$$\in$$\mathbb{R}^{K\times C}$ and $\textbf{T}_0$$\in$$\mathbb{R}^{K\times C}$, respectively.
\begin{equation} 
    \mathcal{L}^{\text{L}}_{\mathrm{reg}} = \text{Cross-Entropy}(\text{Softmax}(\hat{\textbf{t}} \hat{\textbf{T}}_0^{\raisebox{-2pt}{$\scriptstyle\top$}}), \textbf{I}_K),  
\end{equation}
{where $\hat{\textbf{t}}$ and $\hat{\textbf{T}}_0$ are the $\ell_2$ normalized versions of ${\textbf{t}}$ and ${\textbf{T}}_0$ along the channel dimension, respectively, and $\textbf{I}_K$ is a $K$-dimensional identity matrix.}
We use \texttt{``a clean origami of a \textbf{[class]}.''} as $\textbf{P}_0$, 
{
which is 
effective for segmentation\cite{lin2023clip}.}

\vspace{2mm} 
\noindent\textbf{Vision-language regularization.} For improved segmentation capability of textual object queries, we need to preserve joint vision-language alignment at the pixel-level. We incorporate an auxiliary segmentation loss for a pixel-text score map\cite{rao2022denseclip,hummer2023vltseg}. This score map is defined by the cosine-similarity between the visual embeddings $\textbf{x}$ and the text embeddings $\textbf{t}$, computed as $\textbf{S}$$=$$\hat{\textbf{x}}\hat{\textbf{t}}^{\raisebox{-2pt}{$\scriptstyle\top$}}$$\in\mathbb{R}^{hw\times K}$, 
where $\hat{\textbf{x}}$ and $\hat{\textbf{t}}$ are the $\ell_2$ normalized versions of $\textbf{x}$ and $\textbf{t}$ along the channel dimension, respectively. 
The score map $\textbf{S}$ is optimized with a per-pixel cross-entropy loss: 
\begin{equation} 
    \mathcal{L}^{\text{VL}}_{\mathrm{reg}} = \text{Cross-Entropy}(\text{Softmax}(\mathbf{S/\tau}), \mathbf{y}), 
    \label{eq:6}
\end{equation} 
where $\tau$ is a temperature coefficient\cite{he2020momentum}, and $\mathbf{y}$ denotes the ground-truth labels. 

\vspace{2mm}
\noindent\textbf{Vision regularization.} 
In vision transformers \cite{dosovitskiy2020image}, a \texttt{[class]} token captures the global representation of an image \cite{dosovitskiy2020image}. CLIP \cite{radford2021learning} aligns this \texttt{[class]} token with text embedding of a corresponding caption. Therefore, we consider the token as having the preeminent capacity for textual alignment within visual features. To this end, we propose a vision regularization loss $\mathcal{L}^{\text{V}}_{\mathrm{reg}}$ to ensure that the visual backbone preserves its textual alignment at the image-level while learning dense pixel features. Specifically, $\mathcal{L}^{\text{V}}_{\mathrm{reg}}$ enforces the consistency between the \texttt{[class]} token of the training model $\textbf{x}^{\texttt{CLS}}$ and that of the initial visual backbone $\textbf{x}_0^{\texttt{CLS}}$ from a pre-trained VLM:
\begin{equation}
 \mathcal{L}^{\text{V}}_{\mathrm{reg}}=
 \| \mathbf{x}^{\texttt{CLS}} - \mathbf{x}_{\text{0}}^{\texttt{CLS}} \|_2.
\end{equation}

\noindent\textbf{Full objective.} 
The full training objective consists of the segmentation loss $\mathcal{L}_{\mathrm{seg}}$ and the regularization loss $\mathcal{L}_{\mathrm{reg}}=\mathcal{L}^{\text{L}}_{\mathrm{reg}} + \mathcal{L}^{\text{VL}}_{\mathrm{reg}} + \mathcal{L}^{\text{V}}_{\mathrm{reg}}$:
\begin{equation}
    \mathcal{L}_{\mathrm{total}} = \mathcal{L}_{\mathrm{seg}} + \mathcal{L}_{\mathrm{reg}}.
\end{equation}


\section{Experiments}
\subsection{Implementation Details}
\noindent \textbf{Datasets.} 
We evaluate the performance of \texttt{tqdm} under both synthetic-to-real and real-to-real settings. As synthetic datasets, GTA5 \cite{richter2016playing} provides 24,966 images at a resolution of 1914$\times$1052, split into 12,403 for training, 6,382 for validation, and 6,181 for testing. SYNTHIA \cite{ros2016synthia} comprises 6,580 images for training and 2,820 for validation, each at a resolution of 1280$\times$760. As real-world datasets, Cityscapes \cite{cordts2016cityscapes} includes 2,975 images for training and 500 images for validation, with images at 2048$\times$1024 resolution. BDD100K \cite{yu2020bdd100k} consists of 7,000 training and 1,000 validation images, each at 1280$\times$720 resolution. Mapillary \cite{neuhold2017mapillary} offers 18,000 training images and 2,000 validation images, with resolutions varying across the dataset. For simplicity, we abbreviate GTA5, SYNTHIA, Cityscapes, BDD100K, and Mapillary as G, S, C, B, and M, respectively.

\vspace{2mm}
\noindent \textbf{Network architecture.} 
We employ vision transformer-based backbones, initialized with either CLIP\cite{radford2021learning} or EVA02-CLIP\cite{sun2023eva}. The CLIP model incorporates a Vision Transformer-base (ViT-B) backbone\cite{dosovitskiy2020image} with a patch size of 16, while the EVA02-CLIP model utilizes the EVA02-large (EVA02-L) backbone with a patch size of 14. For the pixel decoder, we adopt a multi-scale deformable attention transformer\cite{zhu2020deformable}, which includes $M$$=$$6$ layers, and integrate our text-to-pixel attention layer within it. Regarding the transformer decoder, we follow the default settings outlined in \cite{cheng2022masked}, which consist of $N$$=$$9$ layers with masked attention. The number of textual object queries is set to 19 to align with the number of classes in the Cityscapes dataset \cite{cordts2016cityscapes}. Additionally, the length of the learnable prompt $\textbf{p}$ is set to 8.

\newcolumntype{L}[1]{>{\raggedright\let\newline\\\arraybackslash\hspace{0pt}}m{#1}}
\newcolumntype{C}[1]{>{\centering\let\newline\\\arraybackslash\hspace{0pt}}m{#1}}
\newcolumntype{R}[1]{>{\raggedleft\let\newline\\\arraybackslash\hspace{0pt}}m{#1}}
\def\pd{\phantom{\textsuperscript{\textdagger}}}
\begin{table}[t]
\centering
\caption{
Comparison of mIoU (\%; higher is better) for synthetic-to-real setting (G$\rightarrow$$\{$C, B, M$\}$) and real-to-real setting (C$\rightarrow$$\{$B, M$\}$). \emojiclip,~\emojieva, and \emojidino ~denote initialization with CLIP\cite{radford2021learning}, EVA02-CLIP\cite{sun2023eva}, and DINOv2\cite{oquab2023dinov2} pre-training, respectively. The best and second-best results are \textbf{highlighted} and \underline{underlined}, respectively. Our method is marked in \colorbox{blue}{blue}. The results denoted with $\dagger$ are both trained and tested with an input resolution of $1024\times1024$.
}
\vspace{-1mm}
\renewcommand{\arraystretch}{0.96}
\scalebox{0.92}
{\begin{tabular}{L{2.5cm}|C{1.7cm}|C{1cm}C{1cm}C{1cm}C{1cm}|C{1.05cm}C{1cm}C{1cm}}
\toprule
\multirow{2}{4em}{Method} & \multirow{2}{4em}{Backbone} & \multicolumn{4}{c|}{\textit{\textbf{synthetic-to-real}}} & \multicolumn{3}{c}{\textbf{\textit{real-to-real}}} \\
   &   &  G$\rightarrow$C & G$\rightarrow$B & G$\rightarrow$M & \textbf{Avg.} & C$\rightarrow$B\pd & C$\rightarrow$M\pd & \textbf{Avg.}\pd  \\ 
\midrule
\midrule
SAN-SAW\cite{peng2022semantic} & RN101 & 45.33 & 41.18 & 40.77 & 42.43 & 54.73\pd & 61.27\pd & 58.00\pd  \\ 
WildNet\cite{lee2022wildnet} & RN101 &  45.79 & 41.73 & 47.08 & 44.87 & 47.01\pd & 50.94\pd & 48.98\pd   \\ 
SHADE \cite{zhao2022style} & RN101 & 46.66 & 43.66 & 45.50 & 45.27 & 50.95\pd & 60.67\pd & 55.81\pd  \\ 
TLDR \cite{kim2023texture} & RN101  & 47.58 &  44.88 & 48.80 & 47.09 & - & - & -  \\  
FAMix\cite{fahes2023simple}~\emojiclip & RN101 & 49.47 & 46.40 & 51.97 & 49.28 &-&-&-\\ 
\midrule
SHADE\cite{zhao2023style} & MiT-B5 & 53.27 & \underline{48.19} & 54.99 & 52.15 &-&-&-\\ 
IBAFormer\cite{sun2023ibaformer} & MiT-B5 & \underline{56.34} & \textbf{49.76} & \underline{58.26} & \underline{54.79} &-&-&-\\ 
VLTSeg\cite{hummer2023vltseg}~\emojiclip & ViT-B & 47.50 & 45.70 & 54.30 & 49.17 &-&-&- \\ 
\rowcolor{blue}
\texttt{tqdm} (ours)~\emojiclip   &  ViT-B & \textbf{57.50} & 47.66 & \textbf{59.76} & \textbf{54.97} & \textbf{50.54}\pd & \textbf{65.74}\pd & \textbf{58.14}\pd \\
\midrule
HGFormer\cite{ding2023hgformer} & Swin-L & - & - & - & - & 61.50\pd & 72.10\pd & 66.80\pd \\
VLTSeg\cite{hummer2023vltseg}~\emojieva & EVA02-L & 65.60 & 58.40 & \underline{66.50}	& 63.50 & 64.40\textsuperscript{\textdagger} & \textbf{76.40}\textsuperscript{\textdagger} & \underline{70.40}\textsuperscript{\textdagger}\\
Rein\cite{wei2023stronger}~\raisebox{-0.3mm}{\emojieva} &  EVA02-L & 65.30 & \textbf{60.50} & 64.90 & 63.60 & 64.10\pd & 69.50\pd &66.80\pd\\
Rein\cite{wei2023stronger}~\emojidino & ViT-L & \underline{66.40} & \underline{60.40} & 66.10 & \underline{64.30}  & \textbf{65.00}\pd & 72.30\pd &68.65\pd\\
\rowcolor{blue}
\texttt{tqdm} (ours)~\emojieva  & EVA02-L   & \textbf{68.88} & {59.18} & \textbf{70.10} & \textbf{66.05}  & \underline{64.72}\pd & \underline{76.15}\pd & \textbf{70.44}\pd \\ 

\bottomrule
\end{tabular}}
\label{tab:main_table}

\vspace{-6mm}
\end{table}

\vspace{2mm}
\noindent \textbf{Training.} 
We use the same training configuration for both the CLIP and EVA02-CLIP models. All experiments are conducted using a crop size of $512\times512$, a batch size of 16, and 20k training iterations. Following \cite{cheng2022masked,hummer2023vltseg, wei2023stronger}, we adopt an AdamW \cite{loshchilov2017decoupled} optimizer. 
We set the learning rate at $1\times10^{-5}$ for the synthetic-to-real setting and $1\times10^{-4}$ for the real-to-real setting, with the backbone learning rate reduced by a factor of 0.1. Linear warm-up \cite{goyal2017accurate} is applied over $t_{\text{warm}}$$=$1.5$k$ iterations, followed by a linear decay. We apply standard augmentations for segmentation tasks, including random scaling, random cropping, random flipping, and color jittering. Additionally, we adopt rare class sampling, following \cite{hoyer2022daformer}.

\subsection{Comparison with Previous Methods}

We compare our \texttt{tqdm} with existing DGSS methods. We conduct experiments in two settings: synthetic-to-real (G$\rightarrow$$\{$C, B, M$\}$) and real-to-real (C$\rightarrow$$\{$B, M$\}$), both for the CLIP and EVA02-CLIP models. \cref{tab:main_table} shows that our \texttt{tqdm} generally outperforms the existing methods and achieves state-of-the-art results in both the synthetic-to-real and real-to-real settings. In particular, our approach with the EVA02-CLIP model improves the G$\rightarrow$C benchmark by 2.48 mIoU. More synthetic-to-real setting (\ie, S$\rightarrow$$\{$C, B, M$\}$) results are shown in \cref{sec:C}.

\vspace{-2mm}
\subsection{In-Depth Analysis} \label{sec-5.3}
\begin{figure}[!t]
    \centering
    \begin{subfigure}[t]{0.3\linewidth}
        \centering
        \phantomsubcaption{}
        \label{fig:pr_iou_a}
    \end{subfigure}
    \begin{subfigure}[t]{0.3\linewidth}
        \centering
        \phantomsubcaption{}
        \label{fig:pr_iou_b}
    \end{subfigure}
    \begin{subfigure}[t]{0.3\linewidth}
        \centering
        \phantomsubcaption{}
        \label{fig:pr_iou_c}
    \end{subfigure}
    \includegraphics[width=\linewidth]{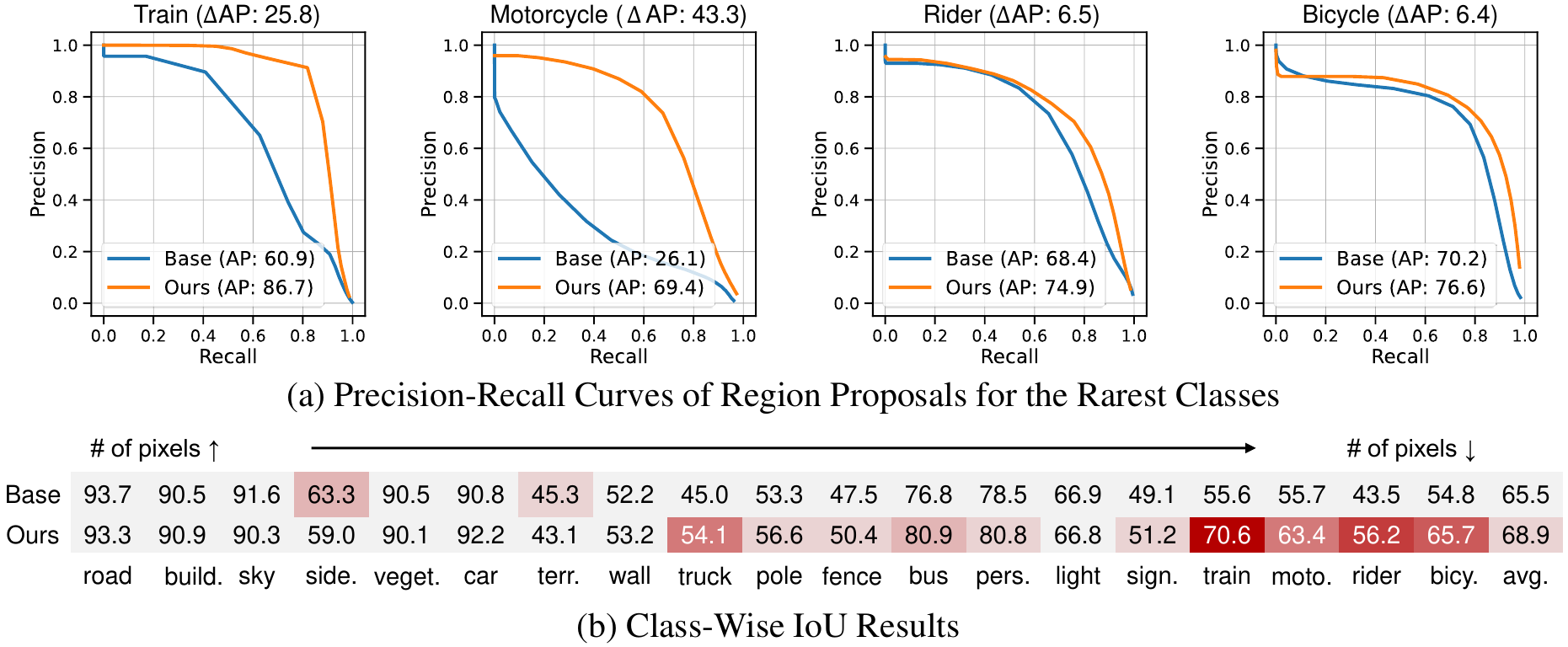}
    \vspace{-15pt}
    \caption{\textbf{Precision-recall curves of region proposals for the rarest classes and class-wise IoU results.} 
    {(a) For the rarest classes (\ie, `train,' `motorcycle,' `rider,' and `bicycle'), our \texttt{tqdm} with textual object query produces more robust region proposals than the baseline with randomly initialized object query. (b) This enhanced robustness leads to the superior DGSS performances of our \texttt{tqdm} for these classes. The red colors visualize the differences in IoU between the baseline and \texttt{tqdm}.}
    }
    \label{fig:pr_iou}
    \vspace{-4mm}
\end{figure}

We design experiments to investigate the factors contributing to the performance improvements achieved by our \texttt{tqdm}. Our analysis focuses on two key aspects: (1) robustness of object query representations for classes of interest, and (2) semantic coherence\footnote{Semantic coherence is a property of vision models in which semantically similar regions in images exhibit similar pixel representations \cite{mukhoti2023open,caron2021emerging}.\label{foot_sc}} of pixel features across domains. We compare 
{our \texttt{tqdm}} model with the baseline Mask2Former\cite{cheng2022masked} model, both initialized with EVA02-CLIP, and trained on GTA5.
The baseline adopts randomly initialized object queries and lacks text-to-pixel attention blocks in its pixel decoder.
For a fair comparison, the baseline employs $K$ object queries with fixed matching, the same as \texttt{tqdm}.

\vspace{2mm}
\noindent \textbf{Robustness of object query representations.} One notable aspect of \texttt{tqdm} is its inherent robustness in object query representations, as we use text embeddings from VLMs as a basis for these queries. 
Given that the role of the initial object query ($\textbf{q}_\textbf{t}^0$) involves localizing region proposals\cite{cheng2022masked} and aggregating pixel information within these proposals to obtain the final object query ($\textbf{q}_\textbf{t}^N$), developing a robust initial object query is crucial for overcoming domain shifts.
Thus, we investigate whether the queries derived from text embeddings produce more robust region proposals compared to {the} randomly initialized {queries}. 

To quantify this robustness, we demonstrate the precision-recall curves of region proposal predictions on G$\rightarrow$C. 
As shown in \cref{fig:pr_iou_a}, for the classes which are rare in the source dataset (\eg, `train,' and `motorcycle')\cite{hoyer2022daformer}, \texttt{tqdm} outperforms the baseline in Average-Precision (AP). 
While the baseline with randomly initialized queries is prone to overfitting for rare classes \cite{li2020analyzing}, our \texttt{tqdm}  
is effective for these classes by encoding domain-invariant representations through the utilization of language information.
{Indeed, the class-wise IoU results in \cref{fig:pr_iou_b} demonstrate notable performance {gains} across multiple classes, with more pronounced gains in rarer ones. These results affirm that the robust region proposals of \texttt{tqdm}, derived from textual object queries, contribute to enhanced final predictions. Further experimental details and results are provided in \cref{sec:D}.}

\begin{figure}[t]
    \centering
    \includegraphics[width=.96\linewidth]{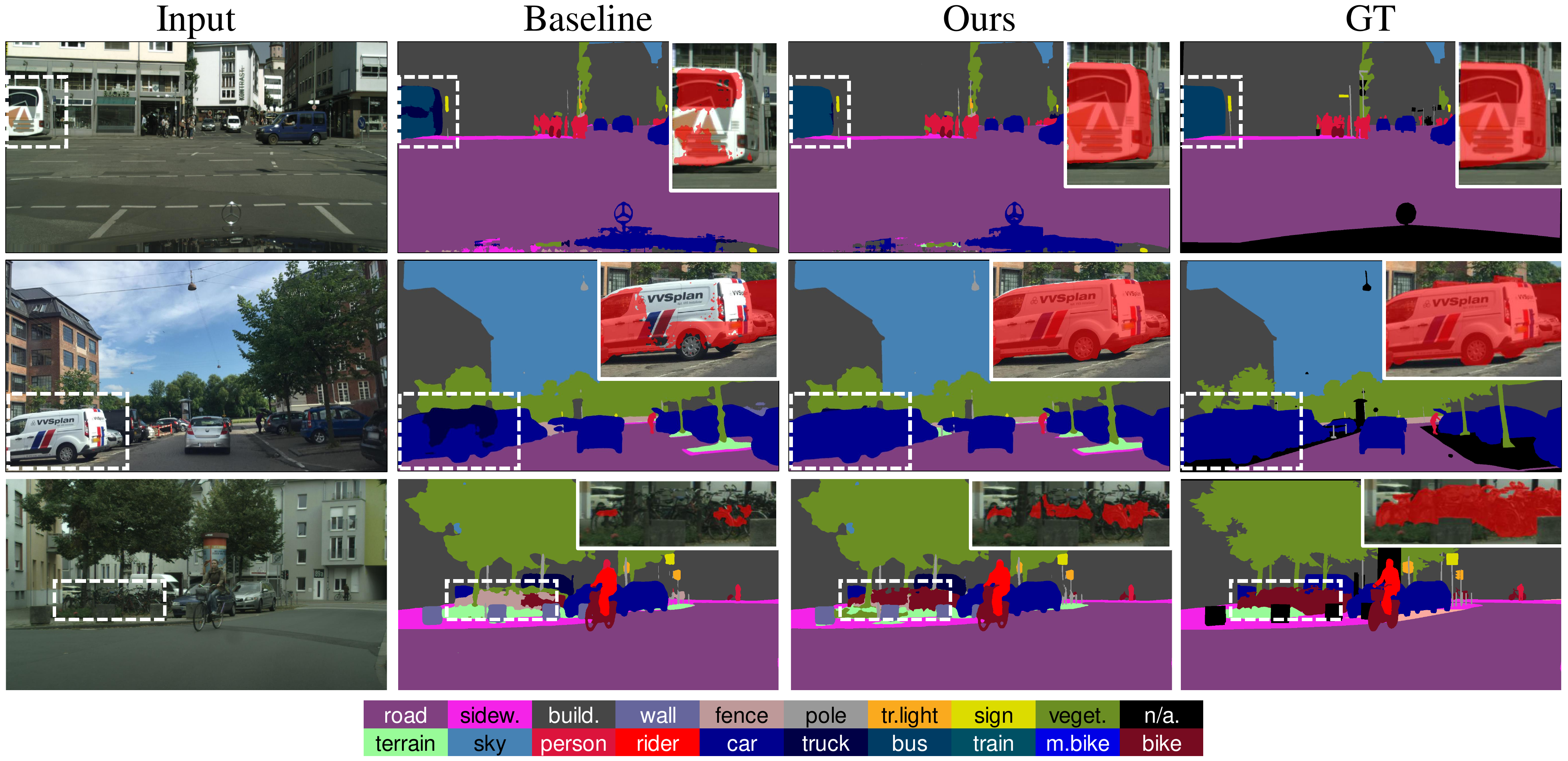}
    \vspace{-5pt}
    \caption{{\textbf{Qualitative results and region proposals by object queries.} Our \texttt{tqdm} provides better prediction results across unseen domains (\ie, C and M) compared to the baseline. We highlight the region proposals generated by object queries with solid lines and the corresponding prediction results with dashed lines. In contrast to the randomly initialized queries of the baseline, our textual object queries lead to more robust region proposals for these classes, resulting in improved predictions.}}
    \label{fig:query_qual}
    \vspace{-4mm}
\end{figure}

Our qualitative results for unseen domains (\ie, C and M), as shown in \cref{fig:query_qual}, also support the idea that robust region proposals lead to enhanced prediction results. 
The \texttt{tqdm} model provides better prediction results with high-quality region proposals, while the baseline produces degraded region proposals, resulting in inferior predictions (refer to white boxes). We provide more qualitative comparisons with other DGSS methods in \cref{sec:E}.

\begin{figure}[t]
    \centering
    \begin{subfigure}[t]{0.3\linewidth}
        \centering
        \phantomsubcaption{}
        \label{fig:tmp_semantic_coherence_a}
    \end{subfigure}
    \begin{subfigure}[t]{0.3\linewidth}
        \centering
        \phantomsubcaption{}
        \label{fig:tmp_semantic_coherence_b}
    \end{subfigure}
    \begin{subfigure}[t]{0.3\linewidth}
        \centering
        \phantomsubcaption{}
        \label{fig:tmp_semantic_coherence_c}
    \end{subfigure}
    \includegraphics[width=0.94\linewidth]{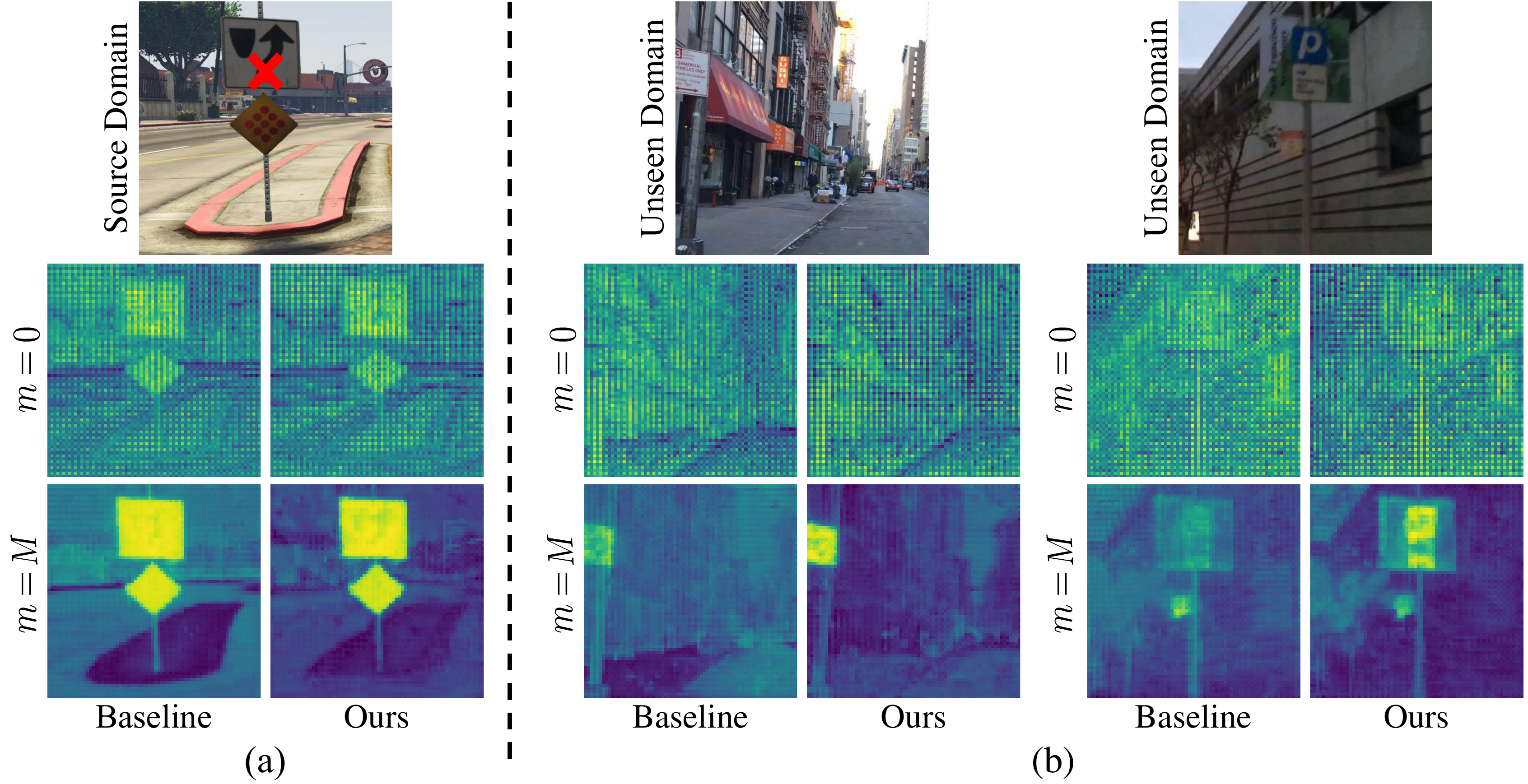}
    \vspace{-2mm}
    \caption{
    {\textbf{Visualization of semantic coherence on pixel features.}} 
    {We compare the semantic coherence of pixel features across source and target domains between \texttt{tqdm} and the baseline. We select a pixel embedding from a source domain image (indicated by \emojicross). Then, we measure the cosine similarities across all pixel embeddings (a) for the source image itself and (b) for unseen domain images, both before ($m$$=$$0$) and after ($m$$=$$M $) processing by the pixel decoder. For the unseen domain images, the \texttt{tqdm} model demonstrates significantly better semantic coherence for the class of interest (\ie, `traffic sign') compared to the baseline after processing by the pixel decoder.}}
    \label{fig:tmp_semantic_coherence}
    \vspace{-4mm}
\end{figure}

\vspace{2mm}
\noindent \textbf{Semantic coherence on pixel features.}
{The other notable aspect of \texttt{tqdm} is its semantic coherence\cite{mukhoti2023open,caron2021emerging} for pixel features across source and target domains. This property is achieved by incorporating text-to-pixel attention within the pixel decoder to enhance the semantic clarity of pixel features. 
{To visualize this property,}
we start by selecting a pixel embedding from a source domain image (indicated by \emojicross ~in \cref{fig:tmp_semantic_coherence}). 
We then measure the cosine similarities across all pixel embeddings for the source image itself and for unseen domain images, both before ($m$$=$$0$) and after ($m$$=$$M$) processing by the pixel decoder. Here, $m$ is the index of pixel decoder layers. Finally, we plot these similarity measurements as heatmaps to visualize the results.}

In the source domain image, both the \texttt{tqdm} and baseline models exhibit comparable levels of semantic coherence before and after processing by the pixel decoder (see \cref{fig:tmp_semantic_coherence_a}). 
{In contrast,}
for the unseen domain images, the \texttt{tqdm} model demonstrates significantly better semantic coherence for the class of interest (\ie, `traffic sign') after processing by the pixel decoder, compared to the baseline (see \cref{fig:tmp_semantic_coherence_b}). These results imply that text-to-pixel attention enhances the semantic clarity of pixel features, consequently leading to the refined pixel features being more effectively grouped by textual object queries, even in unseen domains.

\subsection{Ablation Studies}
\begin{table}[t]
\caption{\textbf{Ablation experiments.} 
    We use the EVA02-L model. The models are trained on GTA5, and evaluated on Cityscapes, BDD100K and Mapillary. 
    The best results are \textbf{highlighted}, and the default setting is marked in \colorbox{blue}{blue}.}
    \vspace{-3mm}
\setlength{\tabcolsep}{2.2pt}
\centering
    \begin{subtable}[h]{0.46\linewidth}
    \centering
    \caption{{\fontsize{8pt}{12pt}\selectfont \textbf{Key Components}}}
    \vspace{-3mm}
    \scalebox{0.9}{
        \begin{tabular}{ccccccc}
        \toprule
        & $\textbf{q}_\text{t}$ & $\textbf{A}_\text{t2p}$
                & C     & B     & M     & Avg.  \\ \hline
        1 & - & - & 66.24 & 57.79 & 68.68 & 64.24 \\
        2 & \checkmark & - & 67.66 & 58.09 & 69.03 & 64.93 \\
        \rowcolor{blue}
        3 & \checkmark & \checkmark & \textbf{68.88}  & \textbf{59.18}  & \textbf{70.10}  & \textbf{66.05}   \\
        \bottomrule
        &&&&&& \\
        \end{tabular}
    }
    \label{tab:ablation-a}
    \end{subtable}
    \hfill
    \begin{subtable}[h]{0.52\linewidth}
    \centering
    \caption{{\fontsize{8pt}{12pt}\selectfont
            \textbf{Regularization Losses}}}
    \vspace{-3mm}
    \scalebox{0.9}{
        \begin{tabular}{cccccccc}
        \toprule
         & $\mathcal{L}^{\mathrm{L}}_\mathrm{reg}$ & $\mathcal{L}^{\mathrm{VL}}_\mathrm{reg}$ & $\mathcal{L}^{\mathrm{V}}_\mathrm{reg}$
                & C     & B     & M     & Avg.  \\ \hline
        1 & - & \checkmark & \checkmark & \textbf{69.34} & 58.01 & 68.98 & 65.44 \\
        2 & \checkmark & - & \checkmark & 65.51 & 55.40 & 68.16 & 63.02 \\
        3 & \checkmark & \checkmark & - & 69.21 & 57.44 & 69.27 & 65.31 \\
        \rowcolor{blue}
        4 & \checkmark & \checkmark & \checkmark & 68.88  & \textbf{59.18}  & \textbf{70.10}  & \textbf{66.05}   \\
        \bottomrule
        \end{tabular}
    }
    \label{tab:ablation-b}
    \end{subtable}
    \\
    \vspace{2mm}
    \begin{subtable}[h]{0.46\linewidth}
    \centering
    \caption{{\fontsize{8pt}{12pt}\selectfont \textbf{Matching Assignment Choice}}}
    \vspace{-3mm}
    \scalebox{0.9}{
        \begin{tabular}{clcccc}
        \toprule
        & Matching & C     & B     & M     & Avg.  \\ \hline
        1 & bipartite & 66.67 & 56.79 & 69.11 &  64.19\\
        \rowcolor{blue}
        2 & fixed & \textbf{68.88}  & \textbf{59.18}  & \textbf{70.10}  & \textbf{66.05}   \\
        \bottomrule
        \end{tabular}
    }
    \label{tab:ablation-c}
    \end{subtable}
    \hfill
    \vspace{2mm}
    \begin{subtable}[h]{0.52\linewidth}
    \centering
    \caption{{\fontsize{8pt}{12pt}\selectfont\textbf{Text Prompt Choice}}}
    \vspace{-3mm}
    \scalebox{0.9}{
        \begin{tabular}{clcccc}
        \toprule
         & Prompt & C     & B     & M     & Avg.  \\ \hline
        1 & fixed template & 68.48 & 57.81 & 69.30 & 65.20 \\
        \rowcolor{blue}
        2 & learnable & \textbf{68.88}  & \textbf{59.18}  & \textbf{70.10}  & \textbf{66.05}   \\
        \bottomrule
        \end{tabular}
    }
    \label{tab:ablation-d}
    \end{subtable}
    
    \vspace{-4mm}
\end{table}
In our ablation experiments, we train the EVA-CLIP model on GTA5 and evaluate it on Cityscapes, BDD100K, and Mapillary. 

\vspace{2mm}
\noindent \textbf{Key components.}
We investigate how the key components contribute to the overall performance of our method. 
We first verify the effectiveness of textual object queries ($\textbf{q}_t$). In \cref{tab:ablation-a}, the model with $\textbf{q}_t$ (row 2) shows better results compared to the model with randomly initialized object query (row 1),
with an average gain of 0.69 mIoU.
Then, we evaluate the contribution of the text-to-pixel attention block ($\textbf{A}_\text{t2p}$), which complements the segmentation capacity of the queries.
$\textbf{A}_\text{t2p}$ further improves the performance by 1.12 mIoU on average, enhancing the semantic clarity of per-pixel embeddings (row 2 and 3).

\vspace{2mm}
\noindent \textbf{Regularization losses.} 
We analyze how each regularization loss contributes to the overall performance. \cref{tab:ablation-b} presents the results under configurations without the language regularization loss ($\mathcal{L}^\mathrm{L}_\mathrm{reg}$), vision-language regularization loss ($\mathcal{L}^\mathrm{VL}_\mathrm{reg}$), and vision regularization loss ($\mathcal{L}^\mathrm{V}_\mathrm{reg}$), respectively. The performance degrades when any of these three regularization losses is excluded, which implies the importance of maintaining robust vision-language alignment. Particularly, $\mathcal{L}^{\mathrm{VL}}_{\mathrm{reg}}$ significantly contributes to the performance by 3.03 mIoU on average (row 2 and 4). This result underlines that the vision-language alignment at the pixel-level is essential for enhancing the efficacy of the textual query-driven framework.

\vspace{2mm}
\noindent \textbf{Matching assignment choice.} 
We adopt fixed matching to ensure that each query represents the semantics of a single class. We compare the fixed matching approach with conventional bipartite matching \cite{cheng2021per,cheng2022masked}.
The model with fixed matching outperforms one with bipartite matching by 1.86 mIoU on average.

\vspace{2mm}
\noindent \textbf{Text prompt choice.} 
We validate the effectiveness of prompt tuning. 
In \cref{tab:ablation-d}, the learnable prompt tuning yields an average of 0.85 mIoU over one that adopts a fixed template prompt (\ie, \texttt{``a clean origami of a [class].''} \cite{lin2023clip}).

\section{Conclusion}
Fine-grained visual features for classes can vary across different domains. This variation presents a challenge for developing models that genuinely understand fundamental semantic knowledge of the classes and generalize well between the domains. To address this challenge in DGSS, we propose utilizing text embeddings from VLMs as object queries within a transformer-based segmentation framework, \ie, textual object queries. Moreover, we introduce a novel framework called \texttt{tqdm} to fully harness the power of textual object queries. Our \texttt{tqdm} is designed to (1) generate textual object queries that fully capture domain invariant semantic information and (2) improve their adaptability in dense predictions through enhancing pixel semantic clarity.
Additionally, we suggest three regularization losses to preserve the robust vision-language alignment of pre-trained VLMs. Our comprehensive experiments demonstrate the effectiveness of textual object queries in recognizing domain-invariant semantic information in DGSS. Notably, \texttt{tqdm} achieves the state-of-the-art performance on multiple DGSS benchmarks, \eg, 68.9 mIoU on GTA5$\rightarrow$Cityscapes, outperforming the prior state-of-the-art method by 2.5 mIoU.

\clearpage  

\section*{Acknowledgements}
We sincerely thank Chanyong Lee and Eunjin Koh for their constructive discussions and support. We also appreciate Junyoung Kim, Chaehyeon Lim and Minkyu Song for providing insightful feedback. This work was supported by the Agency for Defense Development (ADD) grant funded by the Korea government (279002001).

%
%
\bibliographystyle{splncs04}
\bibliography{main}

\begin{thebibliography}{10}
\providecommand{\url}[1]{\texttt{#1}}
\providecommand{\urlprefix}{URL }
\providecommand{\doi}[1]{https://doi.org/#1}

\bibitem{cai2022x}
Cai, Z., Kwon, G., Ravichandran, A., Bas, E., Tu, Z., Bhotika, R., Soatto, S.: {X-DETR: A versatile architecture for instance-wise vision-language tasks}. In: ECCV (2022)

\bibitem{carion2020end}
Carion, N., Massa, F., Synnaeve, G., Usunier, N., Kirillov, A., Zagoruyko, S.: End-to-end object detection with transformers. In: ECCV. Springer (2020)

\bibitem{caron2021emerging}
Caron, M., Touvron, H., Misra, I., J{\'e}gou, H., Mairal, J., Bojanowski, P., Joulin, A.: Emerging properties in self-supervised vision transformers. In: ICCV (2021)

\bibitem{cheng2022masked}
Cheng, B., Misra, I., Schwing, A.G., Kirillov, A., Girdhar, R.: {Masked-attention mask transformer for universal image segmentation}. In: CVPR (2022)

\bibitem{cheng2021per}
Cheng, B., Schwing, A., Kirillov, A.: {Per-pixel classification is not all you need for semantic segmentation}. {NeurIPS}  (2021)

\bibitem{cho2023promptstyler}
Cho, J., Nam, G., Kim, S., Yang, H., Kwak, S.: {PromptStyler: Prompt-driven style generation for source-free domain generalization}. In: ICCV (2023)

\bibitem{cho2023cat}
Cho, S., Shin, H., Hong, S., An, S., Lee, S., Arnab, A., Seo, P.H., Kim, S.: Cat-seg: Cost aggregation for open-vocabulary semantic segmentation. arXiv preprint arXiv:2303.11797  (2023)

\bibitem{choi2021robustnet}
Choi, S., Jung, S., Yun, H., Kim, J.T., Kim, S., Choo, J.: {RobustNet: Improving domain generalization in urban-scene segmentation via instance selective whitening}. In: CVPR (2021)

\bibitem{cordts2016cityscapes}
Cordts, M., Omran, M., Ramos, S., Rehfeld, T., Enzweiler, M., Benenson, R., Franke, U., Roth, S., Schiele, B.: {The cityscapes dataset for semantic urban scene understanding}. In: CVPR (2016)

\bibitem{desai2021virtex}
Desai, K., Johnson, J.: {VirTex: Learning visual representations from textual annotations}. In: CVPR (2021)

\bibitem{ding2023hgformer}
Ding, J., Xue, N., Xia, G.S., Schiele, B., Dai, D.: {HGFormer: Hierarchical grouping transformer for domain heneralized semantic segmentation}. In: CVPR (2023)

\bibitem{dosovitskiy2020image}
Dosovitskiy, A., Beyer, L., Kolesnikov, A., Weissenborn, D., Zhai, X., Unterthiner, T., Dehghani, M., Minderer, M., Heigold, G., Gelly, S., et~al.: An image is worth 16x16 words: Transformers for image recognition at scale. arXiv preprint arXiv:2010.11929  (2020)

\bibitem{fahes2023simple}
Fahes, M., Vu, T.H., Bursuc, A., P{\'e}rez, P., de~Charette, R.: A simple recipe for language-guided domain generalized segmentation. arXiv preprint arXiv:2311.17922  (2023)

\bibitem{fang2023eva02}
Fang, Y., Sun, Q., Wang, X., Huang, T., Wang, X., Cao, Y.: {EVA-02: A visual representation for neon genesis}. arXiv preprint arXiv:2303.11331  (2023)

\bibitem{fang2023eva}
Fang, Y., Wang, W., Xie, B., Sun, Q., Wu, L., Wang, X., Huang, T., Wang, X., Cao, Y.: {EVA: Exploring the limits of masked visual representation learning at scale}. In: CVPR (2023)

\bibitem{goyal2017accurate}
Goyal, P., Doll{\'a}r, P., Girshick, R., Noordhuis, P., Wesolowski, L., Kyrola, A., Tulloch, A., Jia, Y., He, K.: {Accurate, large minibatch sgd: Training ImageNet in 1 hour}. arXiv preprint arXiv:1706.02677  (2017)

\bibitem{he2022masked}
He, K., Chen, X., Xie, S., Li, Y., Doll{\'a}r, P., Girshick, R.: {Masked autoencoders are scalable vision learners}. In: CVPR (2022)

\bibitem{he2020momentum}
He, K., Fan, H., Wu, Y., Xie, S., Girshick, R.: {Momentum contrast for unsupervised visual representation learning}. In: CVPR (2020)

\bibitem{hoyer2022daformer}
Hoyer, L., Dai, D., Van~Gool, L.: {DAFormer: Improving network architectures and training strategies for domain-adaptive semantic segmentation}. In: CVPR (2022)

\bibitem{huang2023sentence}
Huang, Z., Zhou, A., Ling, Z., Cai, M., Wang, H., Lee, Y.J.: {A sentence speaks a thousand images: Domain generalization through distilling CLIP with language guidance}. In: ICCV (2023)

\bibitem{hummer2023vltseg}
H{\"u}mmer, C., Schwonberg, M., Zhong, L., Cao, H., Knoll, A., Gottschalk, H.: {VLTSeg: Simple transfer of CLIP-based vision-language representations for domain generalized semantic segmentation}. arXiv preprint arXiv:2312.02021  (2023)

\bibitem{jia2021scaling}
Jia, C., Yang, Y., Xia, Y., Chen, Y.T., Parekh, Z., Pham, H., Le, Q., Sung, Y.H., Li, Z., Duerig, T.: Scaling up visual and vision-language representation learning with noisy text supervision. In: ICML (2021)

\bibitem{kamath2021mdetr}
Kamath, A., Singh, M., LeCun, Y., Synnaeve, G., Misra, I., Carion, N.: {MDETR -- Modulated detection for end-to-end multi-modal understanding}. In: ICCV (2021)

\bibitem{kim2023texture}
Kim, S., Kim, D.h., Kim, H.: {Texture learning domain randomization for domain generalized segmentation}. ICCV  (2023)

\bibitem{lee2022wildnet}
Lee, S., Seong, H., Lee, S., Kim, E.: {WildNet: Learning domain generalized semantic segmentation from the wild}. In: CVPR (2022)

\bibitem{li2022languagedriven}
Li, B., Weinberger, K.Q., Belongie, S., Koltun, V., Ranftl, R.: Language-driven semantic segmentation. In: ICLR (2022)

\bibitem{li2022dn}
Li, F., Zhang, H., Liu, S., Guo, J., Ni, L.M., Zhang, L.: {DN-DETR: Accelerate detr training by introducing query denoising}. In: CVPR (2022)

\bibitem{li2023transformer}
Li, X., Ding, H., Zhang, W., Yuan, H., Pang, J., Cheng, G., Chen, K., Liu, Z., Loy, C.C.: Transformer-based visual segmentation: A survey. arXiv preprint arXiv:2304.09854  (2023)

\bibitem{li2023clip}
Li, Y., Wang, H., Duan, Y., Li, X.: Clip surgery for better explainability with enhancement in open-vocabulary tasks. arXiv preprint arXiv:2304.05653  (2023)

\bibitem{li2020analyzing}
Li, Z., Kamnitsas, K., Glocker, B.: Analyzing overfitting under class imbalance in neural networks for image segmentation. IEEE transactions on medical imaging  \textbf{40}(3),  1065--1077 (2020)

\bibitem{li2022panoptic}
Li, Z., Wang, W., Xie, E., Yu, Z., Anandkumar, A., Alvarez, J.M., Luo, P., Lu, T.: {Panoptic segformer: Delving deeper into panoptic segmentation with transformers}. In: CVPR (2022)

\bibitem{liang2023open}
Liang, F., Wu, B., Dai, X., Li, K., Zhao, Y., Zhang, H., Zhang, P., Vajda, P., Marculescu, D.: {Open-vocabulary semantic segmentation with mask-adapted CLIP}. In: CVPR (2023)

\bibitem{lin2023clip}
Lin, Y., Chen, M., Wang, W., Wu, B., Li, K., Lin, B., Liu, H., He, X.: {CLIP is also an efficient segmenter: A text-driven approach for weakly supervised semantic segmentation}. In: CVPR (2023)

\bibitem{liu2022dab}
Liu, S., Li, F., Zhang, H., Yang, X., Qi, X., Su, H., Zhu, J., Zhang, L.: {DAB-DETR: Dynamic anchor boxes are better queries for detr}. ICLR  (2022)

\bibitem{liu2023boosting}
Liu, Y., Liu, C., Han, K., Tang, Q., Qin, Z.: {Boosting semantic segmentation from the perspective of explicit class embeddings}. In: ICCV (2023)

\bibitem{loshchilov2017decoupled}
Loshchilov, I., Hutter, F.: {Decoupled weight decay regularization}. ICLR  (2019)

\bibitem{lu2019vilbert}
Lu, J., Batra, D., Parikh, D., Lee, S.: {ViLBERT: Pretraining task-agnostic visiolinguistic representations for vision-and-language tasks}. NeurIPS  (2019)

\bibitem{mangla2022indigo}
Mangla, P., Chandhok, S., Aggarwal, M., Balasubramanian, V.N., Krishnamurthy, B.: {INDIGO: intrinsic multimodality for domain generalization}. arXiv preprint arXiv:2206.05912  (2022)

\bibitem{milletari2016v}
Milletari, F., Navab, N., Ahmadi, S.A.: {V-net: Fully convolutional neural networks for volumetric medical image segmentation}. In: 3DV (2016)

\bibitem{mukhoti2023open}
Mukhoti, J., Lin, T.Y., Poursaeed, O., Wang, R., Shah, A., Torr, P.H., Lim, S.N.: Open vocabulary semantic segmentation with patch aligned contrastive learning. In: CVPR (2023)

\bibitem{neuhold2017mapillary}
Neuhold, G., Ollmann, T., Rota~Bulo, S., Kontschieder, P.: {The mapillary vistas dataset for semantic understanding of street scenes}. In: ICCV (2017)

\bibitem{nguyen2022quality}
Nguyen, T., Ilharco, G., Wortsman, M., Oh, S., Schmidt, L.: {Quality not quantity: On the interaction between dataset design and robustness of CLIP}. NeurIPS  (2022)

\bibitem{oquab2023dinov2}
Oquab, M., Darcet, T., Moutakanni, T., Vo, H., Szafraniec, M., Khalidov, V., Fernandez, P., Haziza, D., Massa, F., El-Nouby, A., et~al.: {DINOv2: Learning robust visual features without supervision}. arXiv preprint arXiv:2304.07193  (2023)

\bibitem{pan2018two}
Pan, X., Luo, P., Shi, J., Tang, X.: {Two at once: Enhancing learning and generalization capacities via ibn-net}. In: ECCV (2018)

\bibitem{pan2019switchable}
Pan, X., Zhan, X., Shi, J., Tang, X., Luo, P.: Switchable whitening for deep representation learning. In: ICCV (2019)

\bibitem{peng2022semantic}
Peng, D., Lei, Y., Hayat, M., Guo, Y., Li, W.: Semantic-aware domain generalized segmentation. In: CVPR (2022)

\bibitem{pham2023combined}
Pham, H., Dai, Z., Ghiasi, G., Kawaguchi, K., Liu, H., Yu, A.W., Yu, J., Chen, Y.T., Luong, M.T., Wu, Y., et~al.: Combined scaling for zero-shot transfer learning. Neurocomputing  (2023)

\bibitem{radford2021learning}
Radford, A., Kim, J.W., Hallacy, C., Ramesh, A., Goh, G., Agarwal, S., Sastry, G., Askell, A., Mishkin, P., Clark, J., et~al.: {Learning transferable visual models from natural language supervision}. In: ICML (2021)

\bibitem{rao2022denseclip}
Rao, Y., Zhao, W., Chen, G., Tang, Y., Zhu, Z., Huang, G., Zhou, J., Lu, J.: {DenseCLIP: Language-guided dense prediction with context-aware prompting}. In: CVPR (2022)

\bibitem{richter2016playing}
Richter, S.R., Vineet, V., Roth, S., Koltun, V.: {Playing for data: Ground truth from computer games}. In: ECCV (2016)

\bibitem{ros2016synthia}
Ros, G., Sellart, L., Materzynska, J., Vazquez, D., Lopez, A.M.: {The SYNTHIA dataset: A large collection of synthetic images for semantic segmentation of urban scenes}. In: CVPR (2016)

\bibitem{sun2023ibaformer}
Sun, Q., Chen, H., Zheng, M., Wu, Z., Felsberg, M., Tang, Y.: {IBAFormer: Intra-batch Attention Transformer for Domain Generalized Semantic Segmentation}. arXiv preprint arXiv:2309.06282  (2023)

\bibitem{sun2023eva}
Sun, Q., Fang, Y., Wu, L., Wang, X., Cao, Y.: Eva-clip: Improved training techniques for clip at scale. arXiv preprint arXiv:2303.15389  (2023)

\bibitem{tan2019lxmert}
Tan, H., Bansal, M.: {LXMERT: Learning cross-modality encoder representations from transformers}. arXiv preprint arXiv:1908.07490  (2019)

\bibitem{vaswani2017attention}
Vaswani, A., Shazeer, N., Parmar, N., Uszkoreit, J., Jones, L., Gomez, A.N., Kaiser, {\L}., Polosukhin, I.: {Attention is all you need}. NeurIPS  (2017)

\bibitem{wei2023stronger}
Wei, Z., Chen, L., Jin, Y., Ma, X., Liu, T., Lin, P., Wang, B., Chen, H., Zheng, J.: Stronger, fewer, \& superior: Harnessing vision foundation models for domain generalized semantic segmentation. arXiv preprint arXiv:2312.04265  (2023)

\bibitem{wortsman2022robust}
Wortsman, M., Ilharco, G., Kim, J.W., Li, M., Kornblith, S., Roelofs, R., Lopes, R.G., Hajishirzi, H., Farhadi, A., Namkoong, H., et~al.: {Robust fine-tuning of zero-shot models}. In: CVPR (2022)

\bibitem{xu2022simple}
Xu, M., Zhang, Z., Wei, F., Lin, Y., Cao, Y., Hu, H., Bai, X.: {A simple baseline for open-vocabulary semantic segmentation with pre-trained vision-language model}. In: ECCV (2022)

\bibitem{yu2020bdd100k}
Yu, F., Chen, H., Wang, X., Xian, W., Chen, Y., Liu, F., Madhavan, V., Darrell, T.: {BDD100k: A diverse driving dataset for heterogeneous multitask learning}. In: CVPR (2020)

\bibitem{yu2022k}
Yu, Q., Wang, H., Qiao, S., Collins, M., Zhu, Y., Adam, H., Yuille, A., Chen, L.C.: {kMaX-DeepLab: k-means mask transformer}. In: ECCV (2022)

\bibitem{yue2019domain}
Yue, X., Zhang, Y., Zhao, S., Sangiovanni-Vincentelli, A., Keutzer, K., Gong, B.: {Domain randomization and pyramid consistency: Simulation-to-real generalization without accessing target domain data}. In: ICCV (2019)

\bibitem{zhang2022segvit}
Zhang, B., Tian, Z., Tang, Q., Chu, X., Wei, X., Shen, C., et~al.: {Segvit: Semantic segmentation with plain vision transformers}. NeurIPS  (2022)

\bibitem{zhang2022dino}
Zhang, H., Li, F., Liu, S., Zhang, L., Su, H., Zhu, J., Ni, L.M., Shum, H.Y.: {DINO: DETR with improved denoising anchor boxes for end-to-end object detection}. arXiv preprint arXiv:2203.03605  (2022)

\bibitem{zhang2023mp}
Zhang, H., Li, F., Xu, H., Huang, S., Liu, S., Ni, L.M., Zhang, L.: {MP-Former: Mask-piloted transformer for image segmentation}. In: CVPR (2023)

\bibitem{zhao2022style}
Zhao, Y., Zhong, Z., Zhao, N., Sebe, N., Lee, G.H.: {Style-hallucinated dual consistency learning for domain generalized semantic segmentation}. In: ECCV (2022)

\bibitem{zhao2023style}
Zhao, Y., Zhong, Z., Zhao, N., Sebe, N., Lee, G.H.: {Style-hallucinated dual consistency learning: A unified framework for visual domain generalization}. IJCV  (2023)

\bibitem{zhou2022extract}
Zhou, C., Loy, C.C., Dai, B.: {Extract free dense labels from CLIP}. In: ECCV (2022)

\bibitem{zhou2022learning}
Zhou, K., Yang, J., Loy, C.C., Liu, Z.: {Learning to prompt for vision-language models}. IJCV  (2022)

\bibitem{zhou2023zegclip}
Zhou, Z., Lei, Y., Zhang, B., Liu, L., Liu, Y.: {ZegCLIP: Towards adapting clip for zero-shot semantic segmentation}. In: CVPR (2023)

\bibitem{zhu2023survey}
Zhu, C., Chen, L.: A survey on open-vocabulary detection and segmentation: Past, present, and future. arXiv preprint arXiv:2307.09220  (2023)

\bibitem{zhu2020deformable}
Zhu, X., Su, W., Lu, L., Li, B., Wang, X., Dai, J.: {Deformable DETR: Deformable transformers for end-to-end object detection}. ICLR  (2020)

\end{thebibliography}

\appendix

\renewcommand\thesection{\Alph{section}}
\pagestyle{empty}
\renewcommand{\theequation}{S\arabic{equation}}
\renewcommand{\thetable}{S\arabic{table}}
\renewcommand{\thefigure}{S\arabic{figure}}
\setcounter{equation}{0}
\setcounter{table}{0}
\setcounter{figure}{0}
\setcounter{page}{1} 

\definecolor{commentcolor}{RGB}{21,101,102}   
\newcommand{\PyComment}[1]{\ttfamily\textcolor{commentcolor}{\# #1}}  
\newcommand{\PyCode}[1]{\ttfamily\textcolor{black}{#1}} 

\definecolor{green}{HTML}{006b3d}
\definecolor{red}{HTML}{b21c1c}
\definecolor{gray}{HTML}{E6E6E6}
\definecolor{blue}{HTML}{e8f0f8}
\definecolor{qrand}{HTML}{2b2e4a}
\definecolor{qtext}{HTML}{e84545}

\definecolor{veget}{RGB}{106,142,34}
\definecolor{terrain}{RGB}{150,251,151}
\definecolor{road}{RGB}{126,64,126}
\definecolor{sidewalk}{RGB}{243,36,231}
\definecolor{person}{RGB}{234,41,55}
\definecolor{rider}{RGB}{255,25,0}
\definecolor{building}{RGB}{68,72,68}

\newcommand{\byeongju}[1]{{\color{red}{#1}}}
\newcommand{\byeonghyun}[1]{{\color{NavyBlue}{#1}}}
\newcommand{\sunghwan}[1]{{\color{orange}{#1}}}

\newcommand{\cmark}{\ding{51}}%
\newcommand{\xmark}{\ding{55}}%

\hypersetup{
  linkcolor = red,
  citecolor  = green,
}

\renewcommand{\figurename}{Figure}

\title{\large Textual Query-Driven Mask Transformer \\ for Domain Generalized Segmentation \\[2.5mm]   {\textmd{Supplementary Material}}}
\author{}
\titlerunning{Textual Query-Driven Mask Transformer}
\authorrunning{B.~Pak et al.}
\institute{}
\maketitle
\vspace{-14mm}


\section*{Appendix}
\vspace{-2mm}
In Appendix, we provide more details and additional experimental results of our proposed \texttt{tqdm}. The sections are organized as follows:
\vspace{-2mm}
\begin{itemize}
   \item \ref{sec:A}. Text Activation in Diverse Domains 
   \item \ref{sec:B}. Details of Motivating Experiment  
   \item \ref{sec:C}. Experiment on SYNTHIA Dataset 
   \item \ref{sec:D}. Details of Region Proposal Experiment 
   \item \ref{sec:E}. More Qualitative Results 
   \item \ref{sec:F}. Qualitative Results on Unseen Game Videos 
   \item \ref{sec:G}. Comparison with Open-Vocabulary Segmentation 
\end{itemize}
\vspace{-7mm}


\section{Text Activation in Diverse Domains\label{sec:A}}
\vspace{-8mm}
\begin{figure}
    \centering
    \includegraphics[width=.93\linewidth]{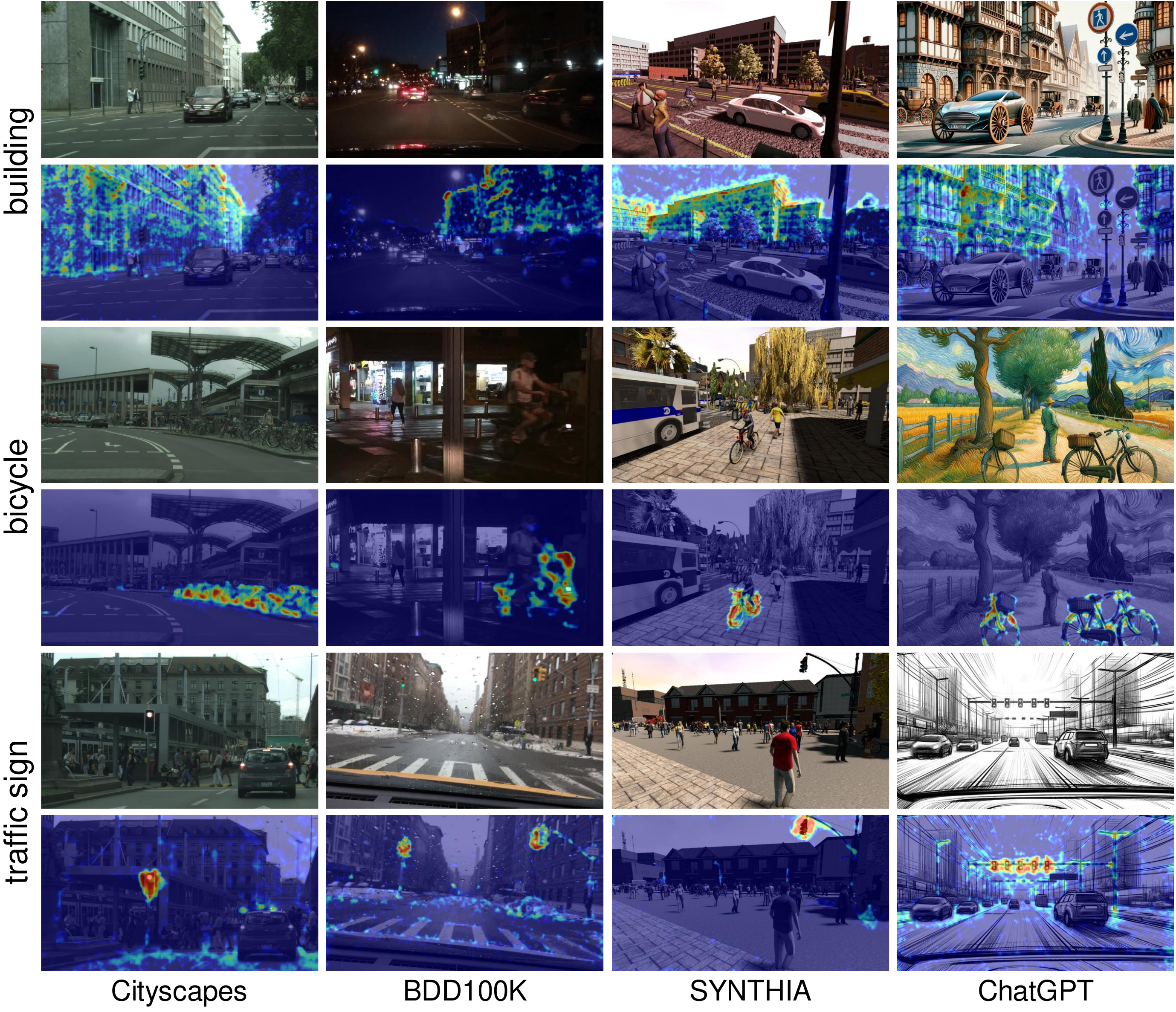}
    \vspace{-2mm}
    \caption{\textbf{Image-text similarity map on diverse domains.} 
    {The text embeddings of the targeted classes (\ie, `building,' `bicycle,' and `traffic sign') are consistently well-activated within the corresponding class regions of images across various domains.}
    }
    \label{fig:text_activation}
\end{figure}

\noindent We find an interesting property of VLMs: the text embedding of a class name is well-aligned with the visual features of the class region across various domains. Specifically, we visualize the image-text similarity map $\textbf{M}$ of a pre-trained VLM\cite{li2023clip,zhou2022extract}, as shown in \cref{fig:text_activation}. {Firstly, we begin by extracting the visual features $\textbf{x}$$\in$$\mathbb{R}^{hw\times C}$ from images across various domains, where $h$ and $w$ are the output resolutions of the image encoder}. In parallel, we obtain the text embeddings $\textbf{t}$$\in$$\mathbb{R}^{K\times C}$ for the names of $K$ classes. Following this, we calculate the similarity map $\hat{\mathbf{x}} \hat{\textbf{t}}^{\raisebox{-2pt}{$\scriptstyle\top$}}$, where $\hat{\textbf{x}}$ and $\hat{\textbf{t}}$ are the $\ell_2$ normalized versions of $\textbf{x}$ and $\textbf{t}$, respectively, along the $C$ dimension.{Lastly, we reshape and resize the resulting similarity map to the original image resolution}. During this process, we also normalize the values using min-max scaling.
The equation to compute $\textbf{M}$ is as follows:
\begin{equation}
    \mathbf{M} = \operatorname{norm}(\operatorname{resize}(\operatorname{reshape}(\hat{\mathbf{x}} \hat{\textbf{t}}^{\raisebox{-2pt}{$\scriptstyle\top$}}))).
\end{equation}
{We adopt the EVA02-CLIP model as a VLM. In \cref{fig:text_activation}, each text embedding (\textit{e.g.}, `bicycle') shows strong activation with the visual features of the corresponding class region across different visual domains. These findings suggest that text embeddings can serve as a reliable basis for domain-invariant pixel grouping.}

\section{Details of Motivating Experiment \label{sec:B}}
\begin{figure}
    \vspace{-7mm}
    \centering
    \includegraphics[width=.96\linewidth]{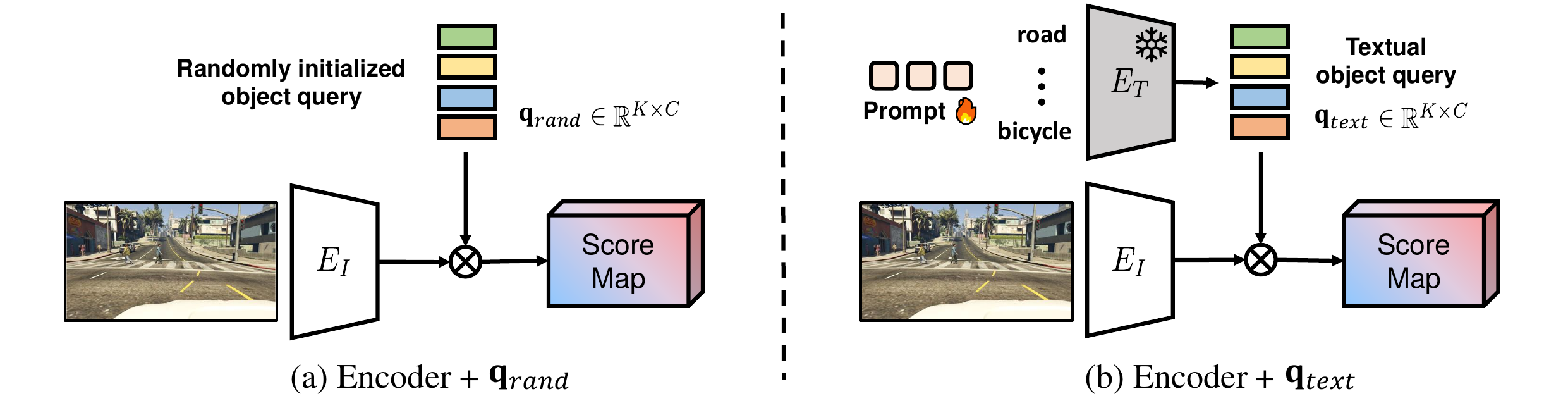}
    \caption{
    {We compare (a) randomly initialized object queries and (b) textual object queries using a simple model architecture, which comprises an image encoder $E_I$ and object queries $\textbf{q}$. For the encoder, we use a ViT-base model with CLIP initialization.}
    }
    \vspace{-1mm}
    \label{supp:motivating}
    \vspace{-4mm}
\end{figure}

{
\noindent In \cref{sec-3}, we demonstrate the superior ability of textual object queries to generalize to unseen domains. As shown in \cref{supp:motivating}, we compare textual object queries with conventional randomly initialized object queries using a simple model architecture. This architecture comprises an image encoder $E_I$ from a VLM and $K$ object queries $\textbf{q}$$\in$$\mathbb{R}^{K\times C}$.  Given {the $\ell_2$ normalized queries $\hat{\textbf{q}}$ and visual embeddings $\hat{\textbf{x}}$$\in$$\mathbb{R}^{hw\times C}$ from $E_I$,} the segmentation logits $\textbf{S}$$=$$\hat{\textbf{x}}\hat{\textbf{q}}^{\raisebox{-2pt}{$\scriptstyle\top$}}$$\in$$\mathbb{R}^{hw\times K}$ are optimized with a per-pixel cross-entropy loss, as described in \cref{eq:6}. In this experiment, we use a CLIP-initialized Vision Transformer-base (ViT-B) backbone with a patch size of 16. The models are trained on GTA5\cite{richter2016playing} or SYNTHIA\cite{ros2016synthia}, with a crop size of 512$\times$512, a batch size of 16, and 5k training iterations. The learning rate is set to $1\times10^{-4}$, and the backbone learning rate is set to $1\times10^{-5}$.
}




\section{Experiment on SYNTHIA Dataset\label{sec:C}}

\begin{table}
\centering
\renewcommand{\arraystretch}{1.0}
\scalebox{0.95}
{\begin{tabular}{L{2.5cm}|C{1cm}C{1cm}C{1cm}C{1cm}}
\toprule
Method   &  S$\rightarrow$C & S$\rightarrow$B & S$\rightarrow$M & \textbf{Avg.}        \\ \midrule
SAN-SAW\cite{peng2022semantic}  & 40.87 & 35.98 & 37.26 & 38.04 \\ 
TLDR \cite{kim2023texture}   & 42.60 &  35.46 & 37.46 & 38.51 \\  
IBAFormer\cite{sun2023ibaformer}  & 50.92 & 44.66 & 50.58 & 48.72 \\ 
VLTSeg\cite{hummer2023vltseg}~\emojieva  & 56.80 & 50.50 & 54.50 & 53.93 \\ 
\rowcolor{blue}
tqdm (ours)~\emojieva &   \textbf{57.99} & \textbf{52.43} & \textbf{54.87} & \textbf{55.10} \\ 
\bottomrule
\end{tabular}}
\vspace{5mm}
\caption{
Comparison of mIoU (\%; higher is better) between DGSS methods trained on S and evaluated on C, B, M. \emojieva ~denotes EVA02-CLIP\cite{sun2023eva} pre-training. The best results are \textbf{highlighted} and our method is marked in \colorbox{blue}{blue}.
}
\vspace{-5mm}
\label{tab:supp_syn}
\end{table}
{
\noindent We conduct an additional experiment in the synthetic-to-real setting (\ie, S$\rightarrow$$\{$C, B, M$\}$), and the results are shown in \cref{tab:supp_syn}. In this experiment, we train on SYNTHIA \cite{ros2016synthia}, and evaluate on Cityscapes \cite{cordts2016cityscapes}, BDD100K \cite{yu2020bdd100k}, and Mapillary \cite{neuhold2017mapillary}. Our \texttt{tqdm} consistently outperforms other DGSS methods across all benchmarks, demonstrating superior synthetic-to-real generalization capability.
}


\section{Details of Region Proposal Experiment \label{sec:D}}

\begin{figure}[t]
    \centering
    \begin{subfigure}[t]{0.49\linewidth}
        \centering
        \phantomsubcaption{}
        \label{fig:supp-pr_a}
    \end{subfigure}
    \begin{subfigure}[t]{0.49\linewidth}
        \centering
        \phantomsubcaption{}
        \label{fig:supp-pr_b}
    \end{subfigure}
    \includegraphics[width=.95\linewidth]{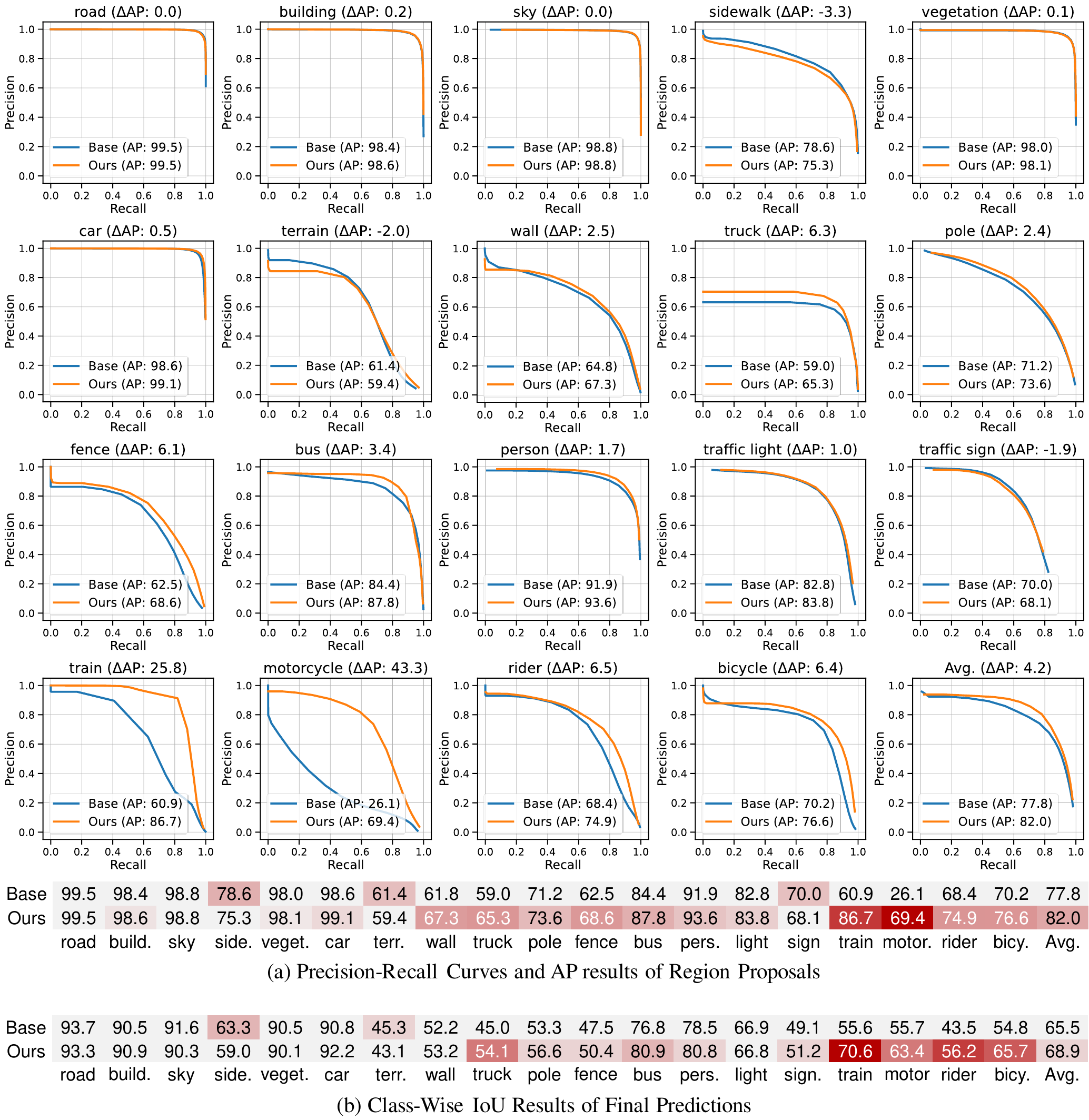}
    \vspace{-1mm}
    \caption{
    {(a) The precision-recall curves and AP results for region proposals. (b) The class-wise IoU results of final predictions. The class-wise trends observed in both tables show a similar pattern.} 
    }
    \label{fig:supp-pr}
    \vspace{-2mm}
\end{figure}

In this section, we provide a detailed explanation of the experiment on the robustness of object query representation discussed in \cref{sec-5.3}, and present further experimental results in \cref{fig:supp-pr}. In \cref{fig:pr_iou}, we compare the region proposal results between our \texttt{tqdm} and the baseline. Given the per-pixel embeddings $\textbf{Z}$$\in$$\mathbb{R}^{H\times W\times D}$ from the pixel decoder, along with the initial object queries $\textbf{q}^0$$\in$$\mathbb{R}^{K\times D}$, region proposals $\textbf{R}$$\in$$\{0,1\}^{H\times W\times K}$ are predicted as follows:

\begin{equation}
    \textbf{R} = \left\{
        \begin{array}{ll}
            1, & \text{ if} ~ \text{sigmoid}( \textbf{Z}{\textbf{q}^0}^{\raisebox{-2pt}{$\scriptstyle\top$}} ) > \raisebox{-0.3mm}{$\theta$} \\
            0, & \text{ otherwise}
        \end{array} \right.
\end{equation}
{where $H$ and $W$ represent the spatial resolutions, $D$ is the channel dimension,  $K$ denotes the number of queries, and \raisebox{-0.1mm}{$\theta$}  is defined as a confidence threshold. By incrementally adjusting $\theta$ from 0.0 to 1.0, we generate precision-recall curves for each region proposal by class, as depicted in \cref{fig:supp-pr_a}. Our \texttt{tqdm} significantly surpasses the baseline in identifying rare classes (i.e., `train,' `motorcycle,' `rider,' and `bicycle'), and achieves marginally better performance across most other classes. Intriguingly, this pattern of enhancement in AP is mirrored in the class-wise IoU results of final predictions, as shown in \cref{fig:supp-pr_b}. These results suggest that the robustness of query representations for semantic regions plays a crucial role in the generalizability of the final mask predictions that stem from these region proposals.}
\clearpage




\section{More Qualitative Results\label{sec:E}}

\begin{figure}
    \centering
    \includegraphics[width=1\linewidth]{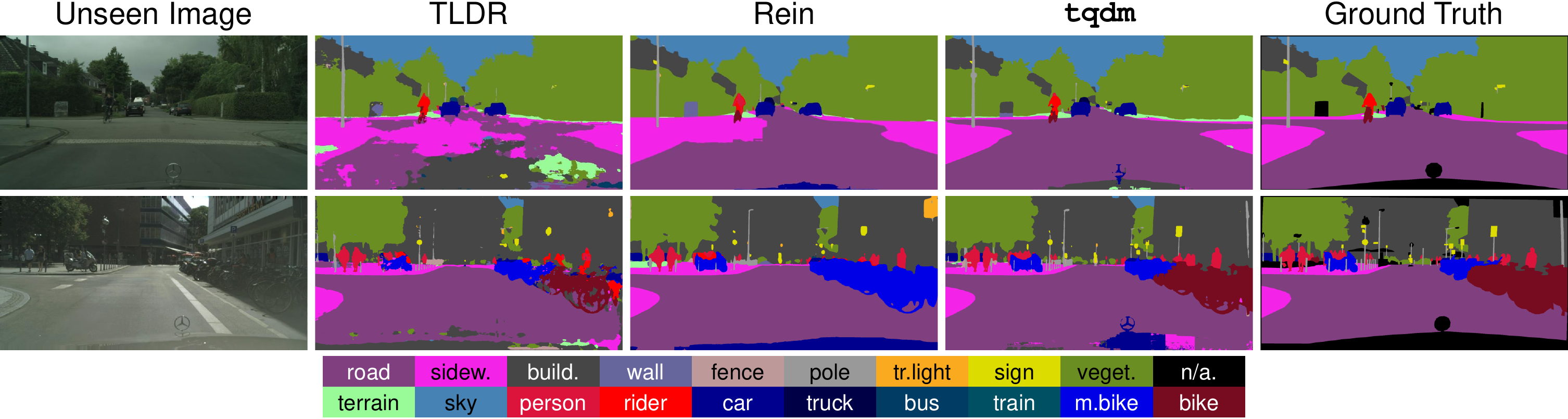}
    \vspace{-4mm}
    \caption{Qualitative results of DGSS methods \cite{kim2023texture,wei2023stronger}, and our \texttt{tqdm} on G$\rightarrow$C.}
    \label{fig:supp-qual-city}
    \vspace{-7mm}
\end{figure}

\begin{figure}
    \centering
    \includegraphics[width=1\linewidth]{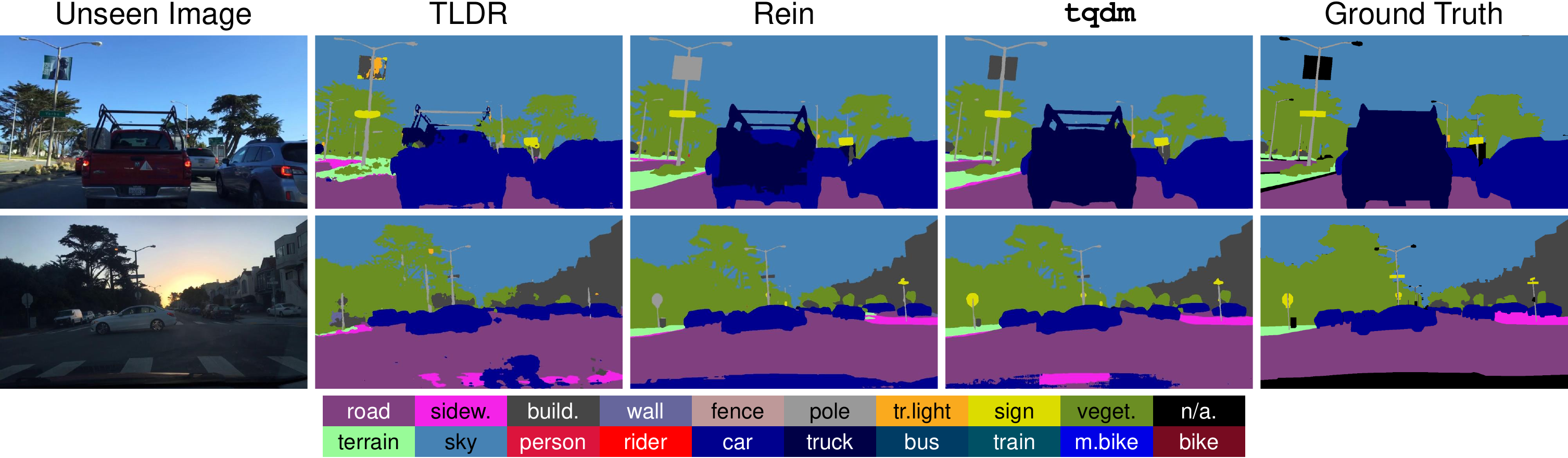}
    \vspace{-4mm}
    \caption{Qualitative results of DGSS methods \cite{kim2023texture,wei2023stronger} and our \texttt{tqdm} on G$\rightarrow$B.}
    \label{fig:supp-qual-bdd}
    \vspace{-7mm}
\end{figure}

\begin{figure}
    \centering
    \includegraphics[width=1\linewidth]{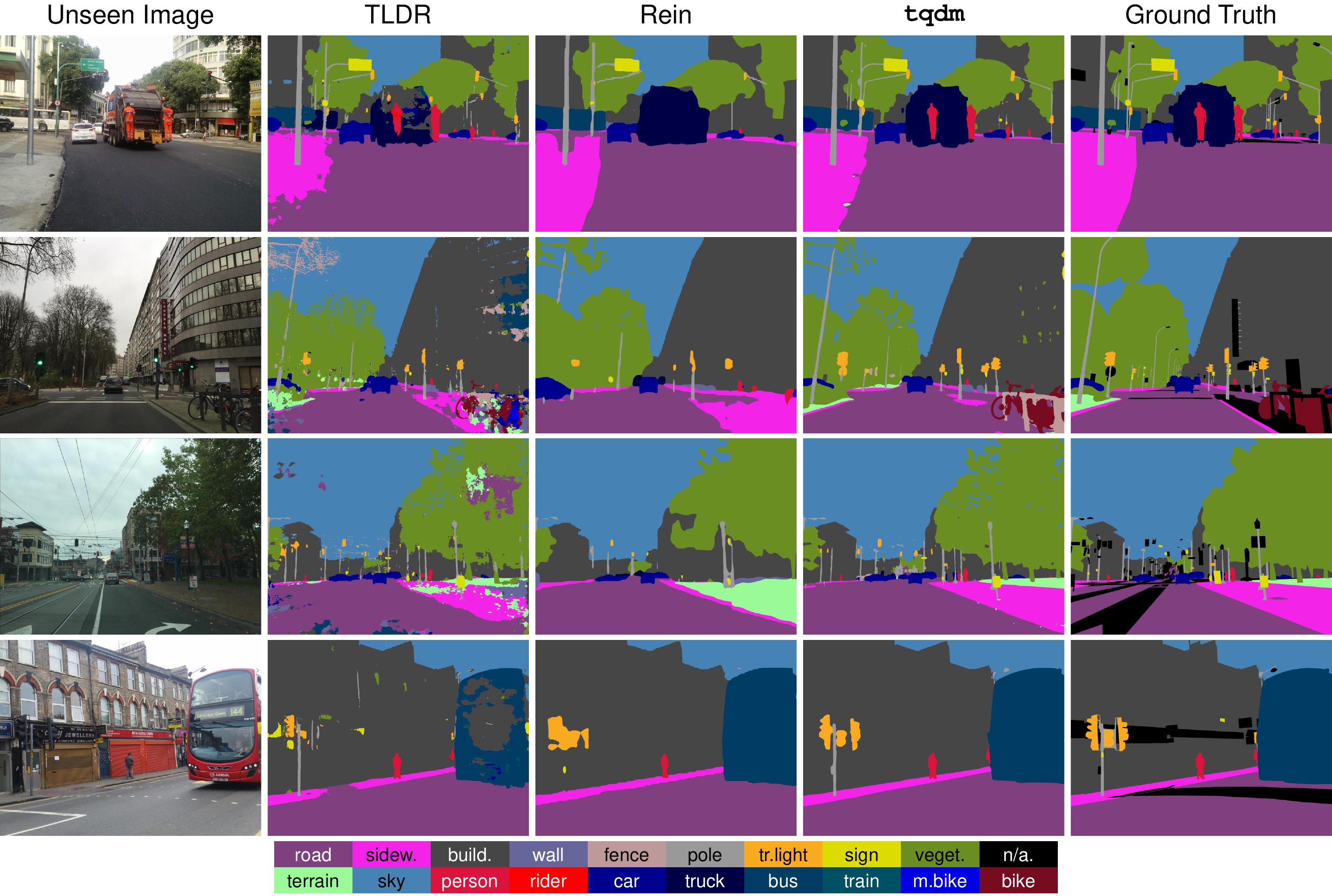}
    \vspace{-4mm}
    \caption{Qualitative results of DGSS methods \cite{kim2023texture,wei2023stronger} and our \texttt{tqdm} on G$\rightarrow$M.}
    \label{fig:supp-qual-map}
\end{figure}

In this section, we provide qualitative comparisons with other DGSS methods \cite{kim2023texture,wei2023stronger}, which demonstrate superior performance using ResNet \cite{kim2023texture} and ViT \cite{wei2023stronger} encoders, respectively. All the models are trained on GTA5 \cite{richter2016playing}. \cref{fig:supp-qual-city,fig:supp-qual-bdd,fig:supp-qual-map} display the qualitative results in the synthetic-to-real setting (G$\rightarrow$$\{$C, B, M$\}$), respectively. Our \texttt{tqdm} yields superior results compared to existing DGSS methods\cite{kim2023texture,wei2023stronger} across various domains.

\begin{figure}[t]
    \centering
    \includegraphics[width=1\linewidth]{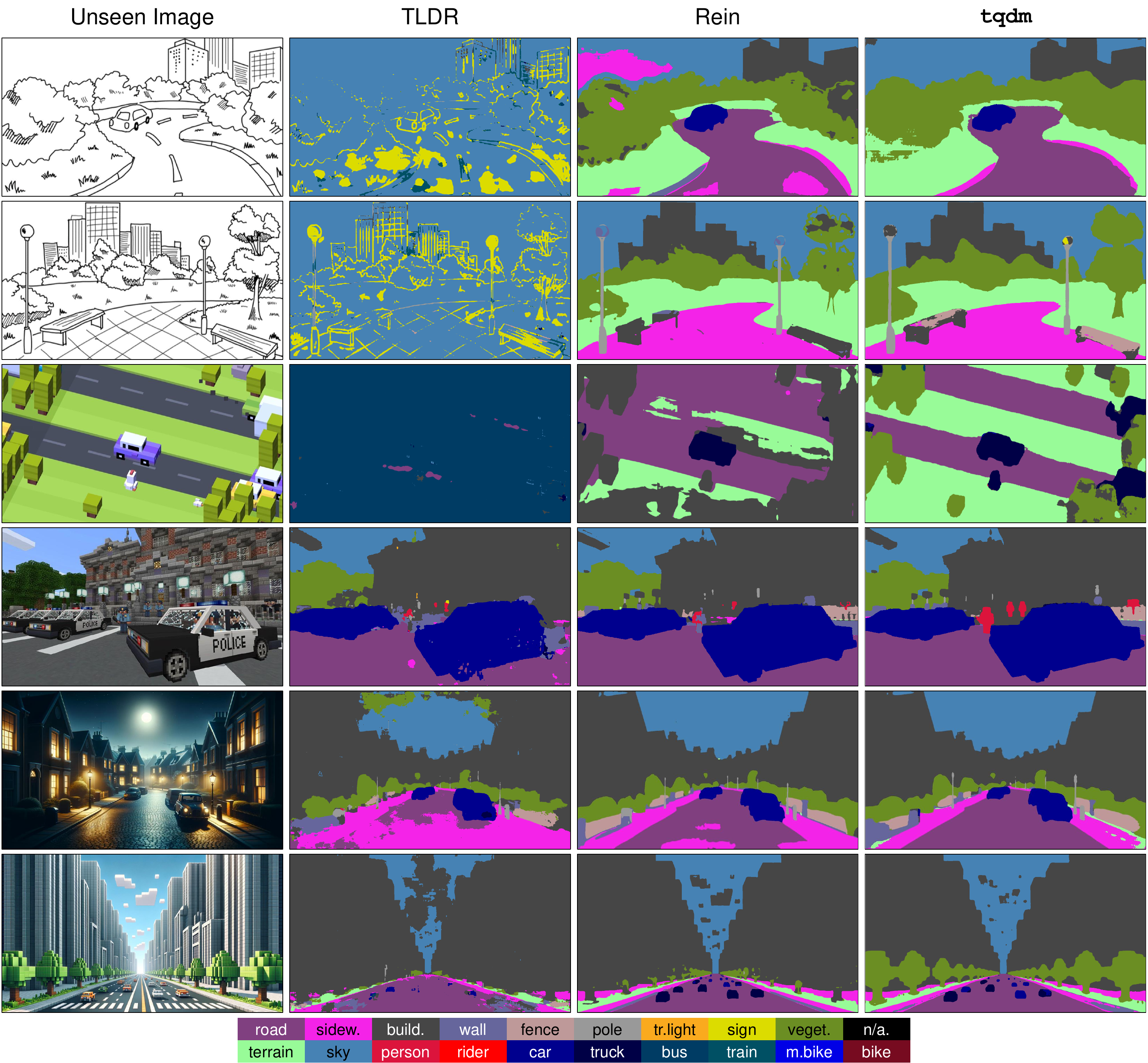}
    \vspace{-4mm}
    \caption{
    {Qualitative results of DGSS methods \cite{kim2023texture,wei2023stronger} and our \texttt{tqdm}, trained on G and evaluated under extreme domain shifts. We present results for hand-drawn images (row 1 and 2), game scene images (row 3 and 4), and images generated by ChatGPT (row 5 and 6).
    }
    }
    \label{fig:supp-qual-gpt}
\end{figure}

{In \cref{fig:supp-qual-gpt}, we further present qualitative comparisons under extreme domain shifts, showcasing results for hand-drawn images (row 1 and 2), game scene images (row 3 and 4), and images generated by ChatGPT (row 5 and 6). Notably, \texttt{tqdm} demonstrates more accurate predictions even under extreme domain shifts, as it is capable of comprehending semantic knowledge.
}





\section{Qualitative Results on Unseen Game Videos}\label{sec:F}
To ensure more reliable results, we perform qualitative comparisons with other DGSS methods \cite{kim2023texture,wei2023stronger} on unseen videos. All the models are trained on GTA5\cite{richter2016playing}. {Our \texttt{tqdm} consistently outperforms the other methods by delivering accurate predictions in unseen videos.} Notably, in both the first and last clips, \texttt{tqdm} effectively identifies trees as the \colorbox{veget}{\textcolor{white}{vegetation}} class and clearly distinguishes the \colorbox{road}{\textcolor{white}{road}} and \colorbox{terrain}{\textcolor{black}{terrain}} classes. Conversely, Rein\cite{wei2023stronger} often misclassifies the background as \colorbox{road}{\textcolor{white}{road}} and trees as \colorbox{building}{\textcolor{white}{building}}. Furthermore, in the second clip, \texttt{tqdm} shows better predictions especially for the \colorbox{person}{\textcolor{white}{person}} class, including the players and the spectators. These results highlight the promising generalization capabilities of our \texttt{tqdm}.

\section{Comparison with Open-Vocabulary Segmentation \label{sec:G}} 

In this section, we compare our \texttt{tqdm} with Open-Vocabulary Segmentation (OVS) approaches \cite{li2022languagedriven,xu2022simple,liang2023open,zhou2023zegclip,cho2023cat}, which also utilize language information from VLMs for segmentation tasks. The fundamental objective of OVS is to empower segmentation models to identify unseen classes during training. 
Our \texttt{tqdm} is designed to generalize across unseen domains for specific targeted classes, while OVS methods aim to segment unseen classes without emphasizing domain shift. This fundamental distinction leads to different philosophies in model design.
\begin{table}
\centering
\renewcommand{\arraystretch}{1.0}
\scalebox{0.92}


{\begin{tabular}{L{2.5cm}|C{1.5cm}|C{3.1cm}|C{1cm}C{1cm}C{1cm}C{1cm}}
\toprule
Method   & Task & Backbone  & G$\rightarrow$C & G$\rightarrow$B & G$\rightarrow$M & \textbf{Avg.}    \\ \midrule
CAT-Seg \cite{cho2023cat}~\emojieva  & OVS  & EVA02-L + Swin-B  & 57.30 & 51.24 & 61.83 & 56.79 \\  
\rowcolor{blue}
tqdm (ours)~\emojieva    & DGSS & EVA02-L & \textbf{68.88} & \textbf{59.18} & \textbf{70.10} & \textbf{66.05} \\ 
\bottomrule
\end{tabular}}
\vspace{2mm}
\caption{
Comparison of mIoU (\%; higher is better) with the state-of-the-art OVS method \cite{cho2023cat} trained on G and evaluated on C, B, M. \emojieva ~denotes EVA-02\cite{fang2023eva,fang2023eva02} pre-training. The best results are \textbf{highlighted} and our method is marked in \colorbox{blue}{blue}.
}
\label{tab:main_ovs}
\end{table}
\vspace{-8mm}

{We conduct a quantitative comparison with the state-of-the-art OVS method, namely CAT-Seg \cite{cho2023cat},  on DGSS benchmarks. CAT-Seg optimizes the image-text similarity map via cost aggregation, and includes partial fine-tuning of the image encoder. For a fair comparison, both models utilize the EVA02-large backbone with EVA02-CLIP initialization and a 512$\times$512 input crop size. As demonstrated in  \cref{tab:main_ovs}, \texttt{tqdm} outperforms the OVS method in DGSS benchmarks (\ie, G$\rightarrow$$\{$C, B, M$\}$). We conclude that the two models exhibit different areas of specialization.}

\clearpage

\end{document}